# Monitoring morphometric drift in lifelong learning segmentation of the spinal cord

**AUTHORS:**

Enamundram Naga Karthik[1,2], Sandrine Bédard[1], Jan Valošek[1,2,3,4], Christoph S. Aigner[5,6], Elise Bannier[7], Josef Bednařík[8,9], Virginie Callot[10,11], Anna Combes[12,13], Armin Curt[14], Gergely David[14,15], Falk Eippert[16], Lynn Farner[14], Michael G Fehlings[17,18], Patrick Freund[14,19,20], Tobias Granberg[21,22], Cristina Granziera[23], RHSCIR Network Imaging Group[24], Ulrike Horn[16], Tomáš Horák[8,9], Suzanne Humphreys[24], Markus Hupp[14], Anne Kerbrat[25,26], Nawal Kinany[27,28], Shannon Kolind[29], Petr Kudlička[9,30], Anna Lebret[14], Lisa Eunyoung Lee[31], Caterina Mainero[32], Allan R. Martin[33], Megan McGrath[13], Govind Nair[34], Kristin P. O'Grady[13], Jiwon Oh[35], Russell Ouellette[21,22], Nikolai Pfender[14], Dario Pfyffer[36,14], Pierre-François Pradat[37], Alexandre Prat[38], Emanuele Pravatà[39,40], Daniel S. Reich[41], Ilaria Ricchi[27,28], Naama Rotem-Kohavi[24], Simon Schading-Sassenhausen[14], Maryam Seif[14,19], Andrew Smith[42], Seth A Smith[13], Grace Sweeney[13], Roger Tam[43], Anthony Traboulsee[44], Constantina Andrada Treaba[32], Charidimos Tsagkas[41,23], Zachary Vavasour[43], Dimitri Van De Ville[27,28], Kenneth Arnold Weber II[45], Sarath Chandar[2,46], Julien Cohen-Adad[1,2,47,48]

**AFFILIATIONS:**

[1] NeuroPoly Lab, Institute of Biomedical Engineering, Polytechnique Montreal, Montreal, QC, Canada
[2] Mila - Quebec AI Institute, Montreal, QC, Canada
[3] Department of Neurosurgery, Faculty of Medicine and Dentistry, Palacký University Olomouc, Olomouc, Czechia
[4] Department of Neurology, Faculty of Medicine and Dentistry, Palacký University Olomouc, Olomouc, Czechia
[5] Physikalisch-Technische Bundesanstalt (PTB), Braunschweig and Berlin, Germany
[6] Max Planck Research Group MR Physics, Max Planck Institute for Human Development, Berlin, Germany
[7] Department of Neuroradiology, Rennes University Hospital, Rennes, France
[8] Department of Neurology, University Hospital Brno, Brno, Czechia
[9] Faculty of Medicine, Masaryk University, Brno, Czechia
[10] Aix-Marseille Univ, CNRS, CRMBM, Marseille, France
[11] APHM, CHU Timone, CEMEREM, Marseille, France
[12] NMR Research Unit, Queen Square Multiple Sclerosis Centre, UCL Queen Square Institute of Neurology, University College London, London, UK
[13] Vanderbilt University Institute of Imaging Science, Vanderbilt University Medical Center, Nashville, USA
[14] Spinal Cord Injury Center, Balgrist University Hospital, University of Zurich, Zurich, Switzerland
[15] Department of Neuro-Urology, Balgrist University Hospital, University of Zurich, Zurich, Switzerland
[16] Max Planck Research Group Pain Perception, Max Planck Institute for Human Cognitive and Brain Sciences, Leipzig, Germany

[17] Division of Neurosurgery and Spine Program, Department of Surgery, Temerty Faculty of Medicine, University of Toronto, Toronto, ON, Canada
[18] Division of Neurosurgery, Krembil Neuroscience Centre, University Health Network, Toronto, ON, Canada
[19] Max Planck Institute for Human Cognitive and Brain Sciences, Leipzig, Germany
[20] Wellcome Trust Centre for Neuroimaging, Queen Square Institute of Neurology, University College London, London, United Kingdom
[21] Department of Neuroradiology, Karolinska University Hospital, Stockholm, Sweden
[22] Department of Clinical Neuroscience, Karolinska Institutet, Stockholm, Sweden
[23] Translational Imaging in Neurology (ThINk), Department of Biomedical Engineering, Faculty of Medicine, Basel, Switzerland
[24] Praxis Spinal Cord Institute, Vancouver, BC, Canada
[25] EMPENN Research Team, IRISA, CNRS-INSERM-INRIA, Rennes Université, Rennes, France
[26] Neurology Department, Rennes University Hospital, Rennes, France
[27] Neuro-X Institute, Ecole Polytechnique Fédérale de Lausanne (EPFL), Geneva, Switzerland
[28] Department of Radiology and Medical Informatics, University of Geneva, Geneva, Switzerland
[29] Division of Neurology, Department of Medicine and the Djavad Mowafaghian Centre for Brain Health, University of British Columbia, Vancouver, BC, Canada
[30] Multimodal and Functional Imaging Laboratory, Central European Institute of Technology, Brno, Czechia
[31] Institute of Medical Science, University of Toronto, Toronto, ON, Canada
[32] Athinoula A. Martinos Center for Biomedical Imaging, Department of Radiology, Massachusetts General Hospital, Charlestown, MA, USA; Harvard Medical School, Boston, MA, USA
[33] Department of Neurosurgery, University of California Davis, Davis, CA, USA
[34] qMRI Core Facility, National Institute of Neurological Disorders and Stroke, National Institutes of Health, Bethesda, MD, USA
[35] Barlo MS Centre, Division of Neurology, Department of Medicine, St. Michael's Hospital, Toronto, Canada
[36] Department of Anesthesiology, Perioperative and Pain Medicine, Stanford University School of Medicine, Palo Alto, CA, USA
[37] Department of Neurology, Pitie-Salpetriere Hospital, Paris, France
[38] Department of Neuroscience, Université de Montréal, Montréal, QC, Canada
[39] Department of Neuroradiology, Neurocenter of Southern Switzerland, Lugano, Switzerland
[40] Department of Neuroscience, Imaging and Clinical Sciences, Università G. d'Annunzio, Chieti, Italy
[41] Translational Neuroradiology Section, National Institute of Neurological Disorders and Stroke, National Institutes of Health, Bethesda, MD, USA
[42] Department of Physical Medicine and Rehabilitation, University of Colorado School of Medicine, Aurora, CO, USA
[43] School of Biomedical Engineering, Department of Radiology, The University of British Columbia, Vancouver, BC, Canada
[44] Department of Medicine, Division of Neurology, University of British Columbia, BC, Canada
[45] Division of Pain Medicine, Department of Anesthesiology, Perioperative and Pain Medicine, Stanford University School of Medicine, Palo Alto, CA, USA
[46] Canada CIFAR AI Chair
[47] Functional Neuroimaging Unit, CRIUGM, Université de Montréal, Montreal, QC, Canada
[48] Centre de Recherche du CHU Sainte-Justine, Université de Montréal, Montréal, QC, Canada

**Corresponding Authors:** naga-karthik.enamundram@polymtl.ca & julien.cohen-adad@polymtl.ca

# ABSTRACT

Morphometric measures derived from spinal cord segmentations can serve as diagnostic and prognostic biomarkers in neurological diseases and injuries affecting the spinal cord. For instance, the spinal cord cross-sectional area can be used to monitor cord atrophy in multiple sclerosis and to characterize compression in degenerative cervical myelopathy. While robust, automatic segmentation methods to a wide variety of contrasts and pathologies have been developed over the past few years, whether their predictions are stable as the model is updated using new datasets has not been assessed. This is particularly important for deriving normative values from healthy participants. In this study, we present a spinal cord segmentation model trained on a multisite (*n=75 sites, 1631 participants*) dataset, including 9 different MRI contrasts and several spinal cord pathologies. We also introduce a lifelong learning framework to automatically monitor the morphometric drift as the model is updated using additional datasets. The framework is triggered by an automatic GitHub Actions workflow every time a new model is created, recording the morphometric values derived from the model's predictions over time. As a *real-world* application of the proposed framework, we employed the spinal cord segmentation model to update a recently-introduced normative database of healthy participants containing commonly used measures of spinal cord morphometry. Results showed that: (i) our model performs well compared to its previous versions and existing pathology-specific models on the lumbar spinal cord, images with severe compression, and in the presence of intramedullary lesions and/or atrophy achieving an average Dice score of 0.95 ± 0.03; (ii) the automatic workflow for monitoring morphometric drift provides a quick feedback loop for developing future segmentation models; and (iii) the scaling factor required to update the database of morphometric measures is nearly constant among slices across the given vertebral levels, showing minimum drift between the current and previous versions of the model monitored by the framework. The code and model are open-source and accessible via Spinal Cord Toolbox v7.0.

**Keywords:** Segmentation, MRI, Spinal Cord, MLOps, Lifelong Learning, Morphometric Drift

# 1. INTRODUCTION

Spinal cord segmentation is relevant for quantifying morphometric changes, such as cord atrophy in multiple sclerosis (MS) (Bautin & Cohen-Adad, 2021; Losseff et al., 1996; Lukas et al., 2013), compression severity in degenerative cervical myelopathy (DCM) (Horáková et al., 2022; Martin et al., 2018), and spared tissue in spinal cord injury (SCI) (Karthik et al., 2024). The development of a robust and accurate spinal cord segmentation tool requires a large sample size which often involves the collaboration of multiple sites and the inclusion of a wide spectrum of MRI scans spanning various spinal cord pathologies, image resolutions, orientations, contrasts, and potential image artifacts. Consequently, obtaining stable morphometric measurements is challenging, as MRI contrasts with different resolutions (and degrees of anisotropy) have varying levels of partial volume effects, leading to subtle shifts in the boundary between the cord and the cerebrospinal fluid (CSF) (Cohen-Adad et al., 2021a; Valošek & Cohen-Adad, 2024). Furthermore, the stability of morphometric measurements is inherently dependent on the version of the segmentation tool and may drift as newer versions are released. This poses a challenge in studies where morphometric measures (e.g., cross-sectional area) are monitored across time.

Previous work in automatic spinal cord segmentation has been limited by a lack of standardization, with models often developed in isolation using different procedures for creating ground truth masks, different model architectures and varying training strategies (M. Chen et al., 2013; De Leener et al., 2014; Gros et al., 2019; Masse-Gignac et al., 2023; Naga Karthik, Valošek, et al., 2025; Nozawa et al., 2023; Tsagkas et al., 2023). Gros et al. (2019) proposed a collection of contrast-specific models (sct_deepseg_sc) trained on healthy controls and MS patients. It uses a convolutional network with 2D kernels, which fails to capture the full spatial context in 3D, resulting in poor performance in DCM and SCI patients with lesions. Masse-Gignac et al. (Masse-Gignac et al., 2023) developed a cascade of two CNNs, trained separately on axial and sagittal T2w scans, for segmenting injured spinal cords, adapting GT masks from sct_deepseg_sc 2D. Nozawa et al. (Nozawa et al., 2023) focused on the segmentation of compressed spinal cords with 2D UNets using transfer learning from DeepLabv3 models (L.-C. Chen et al., 2017). Bédard et al. (Bédard et al., 2025) introduced contrast_agnostic, a 3D model trained on a dataset of healthy participants (Cohen-Adad et al., 2021b), which generalizes across contrasts but struggles to segment pathological cases. The existence of numerous specialized models highlights the lack of standardization in the development of an automatic segmentation pipeline and no continuous learning pipeline exists to monitor or mitigate drift in the segmentation performance of these models over time.

Morphometric measures derived from spinal cord segmentations are highly dependent on the method used (Bédard et al., 2025; Cohen-Adad et al., 2021b) and may drift as the methods evolve. This can lead to inconsistencies in normative values across methods. Moreover, morphometric measures exhibit substantial inter-participant variability driven by factors such as age and sex, which limits sensitivity to subtle changes (Bédard et al., 2024; Bédard &

Cohen-Adad, 2022; Labounek et al., 2024; Papinutto et al., 2020; Taso et al., 2016; Valošek et al., 2024). One approach to mitigate this variability is to compare them with morphometrics obtained from healthy controls (Bédard et al., 2024; Horáková et al., 2022; Kato et al., 2012; Labounek et al., 2024; Valošek et al., 2024). These normalization techniques assume that the morphometrics from new participants are computed using the same method as the normative reference (Valošek et al., 2024)—an assumption that no longer holds as segmentation methods are iteratively improved upon, highlighting the need for population databases to evolve alongside segmentation techniques.

Given that the aforementioned tools only target a limited set of pathologies, often with few MRI contrasts, there is great value in unifying their specialized analyses into a single model which could work with a substantially larger, cumulative, training set. With segmentation frameworks such as nnUNetV2 (Isensee et al., 2021), which has been widely adopted by the medical imaging community due to its robustness and generalization to several modalities and neural network architectures (Isensee et al., 2024), achieving this objective is now possible. In addition, a standardized training strategy to continuously update models over time, monitor performance drift between various model updates, and manage model retraining would streamline these approaches substantially. Such a lifelong learning framework (Agirre et al., 2021; Liu & Mazumder, 2021; Prapas et al., 2021) ensures that the model remains robust to shifts in the data distribution and continually refine their segmentation performance across the diverse set of contrasts and pathologies (Karthik et al., 2022).

To address these challenges, our study contributes the following:

1. An automatic spinal cord segmentation model trained on a multi-site dataset gathered from 75 sites worldwide. This dataset consisted of 9 different MRI contrasts spanning a wide range of image resolutions, including pathologies such as MS (with different phenotypes), traumatic SCI (acute and chronic) and non-traumatic SCI (DCM and ischemic SCI).

2. A lifelong learning framework for developing models to segment new contrasts and pathologies over time. This framework also presents an automatic workflow capable of monitoring the drift in the spinal cord morphometrics across various versions of the models using GitHub Actions.

3. Validation of the lifelong learning framework to update a normative database of spinal cord morphometric measures (Valošek et al., 2024).

The proposed spinal cord segmentation model and normative database are open-source and integrated into the Spinal Cord Toolbox (SCT) (De Leener et al., 2017), accessible as of v7.0.

# 2. MATERIALS AND METHODS

## 2.1. Data curation

### 2.1.1. Data and participants

Our “real-world” dataset contains data from 75 sites and 1,631 participants, including healthy participants (*n=428*), people with degenerative cervical myelopathy (DCM; *n=359*), spinal cord injury (SCI; *n=286*), MS or suspected MS (*n=164*), amyotrophic lateral sclerosis (ALS; *n=13*), neuromyelitis optica spectrum disorder (NMOSD; *n=10*), and syringomyelia (SYR; *n=1*). The MS cohort spanned different phenotypes, ranging from preclinical MS stage (i.e., radiologically isolated syndrome, RIS; *n=61*) to clinically definite MS, including relapsing-remitting MS (RRMS; *n=249*), and primary progressive MS (PPMS; *n=60*). Within the SCI cohort, the images spanned various phases and lesion etiologies of the injury, namely traumatic (*n=171; intermediate and chronic*), acute traumatic (pre-decompression) SCI (*n=95*), ischemic (*n=13*), hemorrhagic (*n=5*), and unknown (*n=2*) lesions. A single participant may contribute one or more different sequences, depending on the site, resulting in a total of 3,453 images (3D volumes[1]). The study included 9 different contrasts, namely, T1-weighted (T1w; $n_{vol.}=318$), T2-weighted (T2w; $n_{vol.}=1377$), T2*-weighted (T2*w; $n_{vol.}=499$), diffusion-weighted (DWI; $n_{vol.}=243$), gradient-echo sequence with (MT-on; $n_{vol.}=248$) and without (GRE-T1w; $n_{vol.}= 243$) magnetization transfer pulse, phase-sensitive inversion recovery (PSIR; $n_{vol.}=333$), short tau inversion recovery (STIR; $n_{vol.}=89$), and MP2RAGE UNIT1 ($n_{vol.}=103$). The images could cover any of the cervical, thoracic and lumbar spinal regions (i.e. the model was trained on chunks containing either of those regions). Whole-spine scans covering all regions are not used for training. Spatial resolutions included isotropic (0.8 mm to 1 mm), anisotropic axially-oriented (in-plane resolution: 0.29 mm to 1 mm; slice thickness: 1 mm to 9.3 mm) and sagittally-oriented (in-plane resolution: 0.28 mm to 1 mm; slice thickness: 0.8 mm to 4.83 mm) images. Images were acquired at 1T, 1.5T, 3T, and 7T on various scanner manufacturers (Siemens, Philips and GE). Figure 1 shows the overall summary of the dataset and **Table S1** provides more details on the distribution of image resolutions for each contrast.

[1] Note that “images” and “volumes” are used interchangeably, both referring to 3D MRI scans.

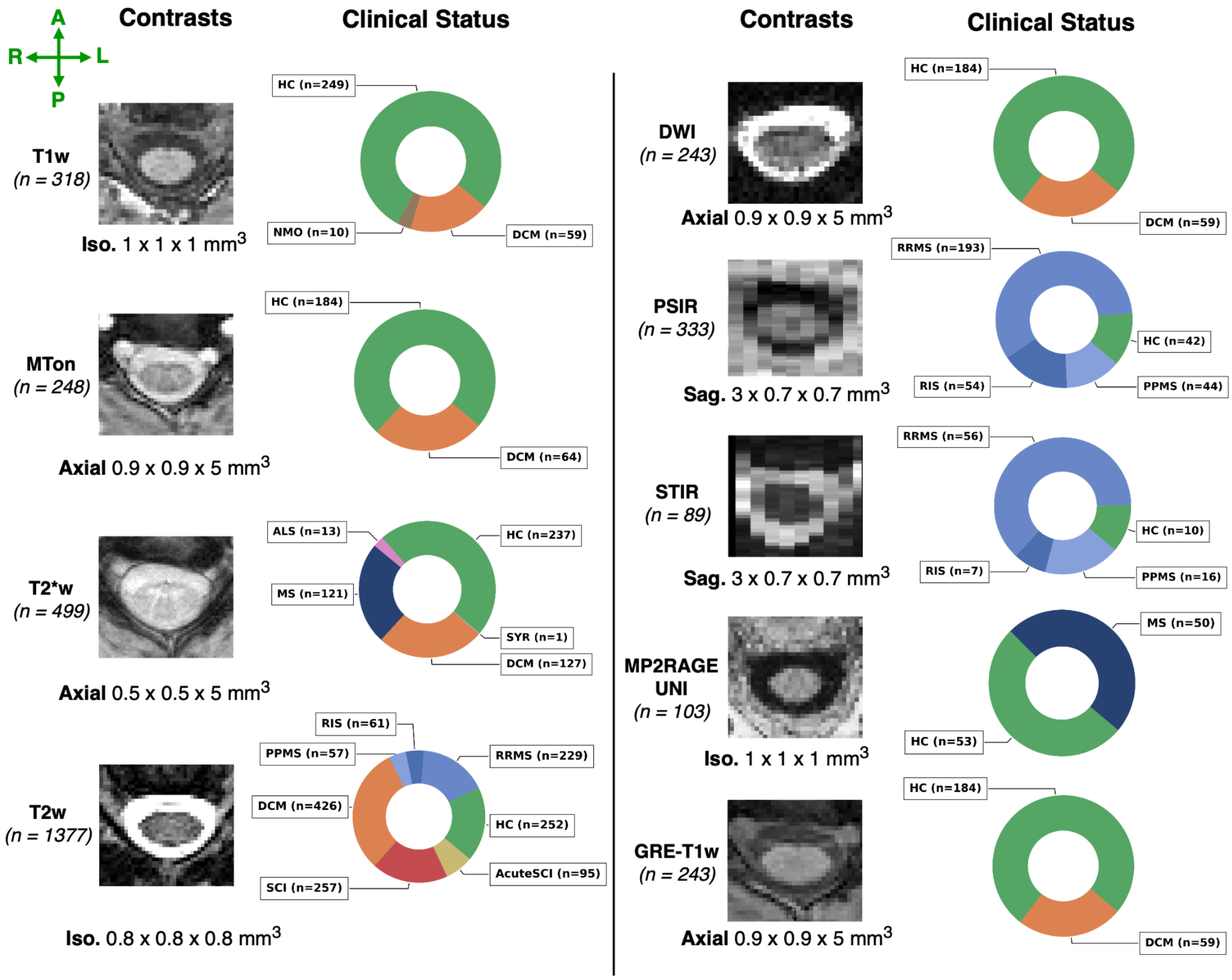

Figure 1 ***Overview of the dataset and image characteristics**. Representative axial slices of 9 contrasts and the total of images used for each contrast in brackets, the orientation (axial/sagittal) along with the median resolution of images. The respective doughnut chart illustrates the proportion of clinical status among the scanned participants, including healthy controls (HC), patients with radiologically isolated syndrome (RIS), patients with multiple sclerosis (MS) and their different phenotypes, including primary progressive (PPMS) and relapsing-remitting (RRMS), patients with amyotrophic lateral sclerosis (ALS), patients with neuromyelitis optica spectrum disorder (NMOSD), pre-decompression acute traumatic SCI (AcuteSCI), post-decompression traumatic spinal cord injury (SCI), degenerative cervical myelopathy (DCM), and syringomyelia (SYR; not shown). Labels indicate the phenotype associated with the patient, with their respective colors shared across contrast sets.*

### 2.1.2. Generating ground truth masks

We used the GT masks in the spine-generic multi-subject database, generated using the same preprocessing procedure from our previous work (Bédard et al., 2025). For the newly obtained datasets, we initially performed a quality control (QC) using `sct_qc`, SCT's visual QC tool (Valošek & Cohen-Adad, 2024). Four experienced raters (ENK, SB, JV, JCA) qualitatively assessed the image-GT pairs and flagged images with motion artifacts and poor signal quality to be excluded from training. In cases where the GT masks were under- or over-segmented (e.g., due to the lower contrast at the spinal cord-cerebrospinal fluid boundary or due to the presence of cord compression), the GT masks were recreated using a combination of the contrast-agnostic model (Bédard et al., 2025) and manual corrections. In datasets with severe deformations to the spinal cord anatomy (e.g., SCI and DCM), a pathology-specific model, SCIseg (Karthik et al., 2024; Naga Karthik, Valošek, et al., 2025) was used instead, followed by manual corrections by JV and ENK. In pathologies involving intramedullary lesions (e.g., MS, SCI, and DCM), lesions were considered part of the spinal cord and included in the GT masks. Manual intervention was required in approximately 9% of the scans (~311 images out of 3,453 in total). All GT masks were binarized using a threshold of 0.5 prior to preprocessing and training to ensure uniformity.

For each site, the data were split participant-wise following an 80%-20% train-test split ratio, ensuring that participants with multiple scans (or multiple sessions), were included either in the training set or the testing set (mutually exclusive). This ensures that no data leakage between train and test splits could occur. After pooling the training and testing data from each participant and each site, the aggregated dataset included 2,945 training and 508 testing images.

## 2.2. Training protocol

### 2.2.1. Preprocessing and data augmentation

We chose the nnUNet framework for training our spinal cord segmentation model as it easily allows future retraining of the model with new contrasts and pathologies and can also be readily integrated into existing open-source packages such as SCT (De Leener et al., 2017; *SlicerNNUnet: 3D Slicer nnUNet Integration to Streamline Usage for nnUNet Based AI Extensions*, n.d.), facilitating broader use by the spinal cord imaging community.

All images and GT masks were re-oriented to right-posterior-inferior (RPI). The median resolution of images in the training set was [0.9 x 0.7 x 1] $mm^3$ and the median shape was [96 x 320 x 318]. Images were resampled to the median resolution using spline interpolation (`order`=3), and GT masks were resampled using linear interpolation (`order`=1). The patch size was set to [64 x 224 x 160]. Standard data augmentation transforms in the nnUNet pipeline, being randomly applied, were predefined with a probability (*p*) and called in the following order: affine transformation (rotation and scaling; *p=0.2*), Gaussian noise addition *(p=0.1)*, Gaussian smoothing *(p=0.2)*, image brightness augmentation (p=0.15), simulation of low resolution with downsampling and upsampling factors sampled uniformly from *[0.5, 1.0]* *(p=0.25)*, gamma correction *(p=0.1*), mirroring transform across all axes. The versatility of these transforms is evident from nnUNet's success on a diverse range of segmentation tasks. Hence, we decided to keep the predefined set of augmentation transforms. Part of the decision for transformations to be applied also stems from empirical evidence from our prior work (Bédard et al., 2025). Lastly, all images were normalized using *z*-score normalization.

#### 2.2.2. Network architecture and training hyperparameters

The network architecture is based on the U-Net with 6 layers in the encoder, starting with 32 feature maps at the initial layer and ending with 320 feature maps at the bottleneck (i.e. 32→64→128→256→320→320) and 5 layers in the decoder with skip-connections at each successive resolution. Each layer in the encoder and the decoder consists of a stack of two convolutional blocks, where each block contains a series of convolutional (with 3x3x3 kernels), instance normalization and LeakyReLU activation layers (with slope=0.1). No dropout layers were used. At each layer in the decoder, deep supervision (Dou et al., 2017) was also used, where auxiliary losses from the feature maps at each upsampling resolution are added to the final loss. The network was trained with a combination of Dice (Milletari et al., 2016) and cross-entropy losses with equal weights. The model was trained using 5-fold cross-validation for 1000 epochs, a batch size of two, and with the stochastic gradient descent (SGD) optimizer and a polynomial learning rate scheduler. Checkpoints were saved based on the exponential moving average of Dice score on the validation set computed at the end of each epoch. **Table S3** contains the values for each hyperparameter mentioned above. All experiments were run on a single 48 GB NVIDIA A6000 GPU.

### 2.3. Lifelong learning for morphometric drift monitoring

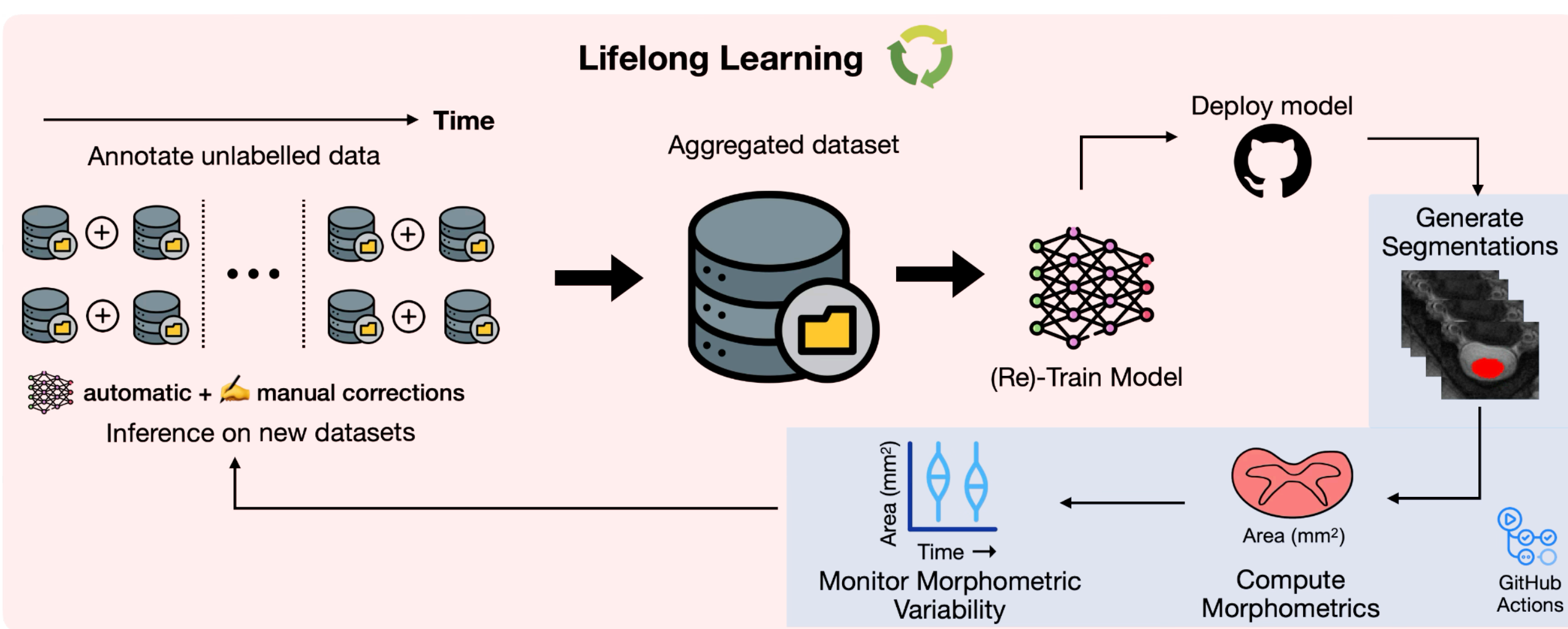

Figure 2 ***Overview of the lifelong learning strategy for continuous training of spinal cord segmentation models.*** *Unlabelled data containing various contrasts and pathologies, gathered from multiple sites worldwide, are segmented automatically with an existing state-of-the-art model and undergo visual quality control for inconsistencies in segmentations, excluding data with artifacts. Labelled datasets are aggregated to train the spinal cord segmentation model. Post-training, the model is deployed as an official release, triggering an automatic GitHub Actions workflow that generates the segmentations, computes the morphometrics, and actively monitors the drift in the morphometric variability between the current version of the model and the previously released versions (automated tasks shown in the blue box). As new data arrive, the process is repeated, enabling continuous (re)training of the models to segment a diverse set of contrasts and pathologies.*

We take an *MLOps* (Alla & Adari, 2020; Tabassam, 2023; Treveil et al., 2020) approach to propose our lifelong learning framework for monitoring morphometric drift across various versions of the model (Figure 2). Once the segmentation model is trained, we deploy the model as an official release on GitHub[2]. The release triggers an automatic GitHub Actions workflow that: (i) downloads the publicly available dataset, (ii) runs the morphometric analysis, (iii) generates the plots quantifying the drift in the performance between the current and previous versions of the model, and (iv) updates the GitHub release assets by uploading the plots and the morphometric values. It is worth emphasizing that all the above steps are performed automatically once a model is released, thus facilitating model development through continuous integration and continuous deployment (CI/CD) (see Figure 3 for pseudocode of the workflow). A key element in the automated workflow is a public test set, which must be 'frozen' to ensure newer models are evaluated fairly against earlier versions. Accordingly, we used the test set from our previous study[3] (Bédard et al., 2025), consisting of 49 healthy participants with 6 contrasts each (T1w, T2w, T2*w, DWI, MT-on, GRE-T1w). Using this data, we computed the spinal cord cross-sectional area (CSA) to monitor morphometric variability. More importantly, monitoring performance drift among models on publicly-available participant data avoids data privacy issues when running the morphometric analysis on the cloud using GitHub Actions workflows. Furthermore, running this task after each model finishes training ensures that the deployed model does not drift too much from the stable version (Bédard et al., 2025). We can then use the current version of the model (which is now the new state-of-the-art) to annotate existing or new unlabelled datasets (arriving in the future), perform QC, add them to growing collection of datasets, and retrain the next version of the model, closing the loop for a continuous learning strategy. Note that this differs from the classical approach to lifelong/continual learning, where it is assumed that access to previously available data is constrained or unavailable (Sodhani et al., 2022), as our new models have unrestrained access to all prior data.

---

[2] https://github.com/sct-pipeline/contrast-agnostic-softseg-spinalcord/releases/tag/v3.0

[3] https://github.com/sct-pipeline/contrast-agnostic-softseg-spinalcord/releases/tag/v2.0

Our choice of using GitHub Actions workflow stems from the ease of accessibility of previous spinal cord segmentation models in SCT (De Leener et al., 2017). When a new model is released on GitHub, it can be easily downloaded using the command `sct_deepseg spinalcord -install -custom-url <release-url>` without having to install any model-specific packages. As a result, the GitHub Actions workflow is simply tasked with installing SCT and running the above-mentioned command for computing morphometrics across various models (accessible via the URL of their releases).

**Algorithm 1** Pseudocode for Monitoring Morphometric Drift

```
name: Run morphometric analysis
on:
  release:
    types: [published]
jobs:
  # job 1: Download the dataset hosted on git-annex
  download_dataset: # define name for the job
    steps:
        # steps performed in the job
        - name: Install git-annex
        - name: Download test data using git-annex
        - name: Cache downloaded dataset

  # job 2: Compute morphometrics
  compute_csa:
    needs: download_dataset # requires previous job to finish
    steps:
        - name: Restore cached dataset
        - name: Install Spinal Cord Toolbox
        - name: Run morphometric analysis on test subset

  # job 3: Generate plots
  generate_plots:
    needs: compute_csa
    steps:
        - name: Generate morphometric drift plots
        - name: Upload plots to GitHub release
```

Figure 3 *Pseudocode of the automatic workflow for monitoring morphometric drift after deploying the segmentation model. The workflow is divided into three jobs: (1) downloading the dataset from* `git-annex`*, (2) running morphometric analysis (computing CSA) across the test set, and (3) generating plots to monitor drift in morphometric variability and updating the GitHub release with the plots. Note that job #2 is parallelized across several GitHub runners on the cloud, where each runner processes a subset of the test set for computational efficiency.*

## 2.4. Validation protocol

### 2.4.1. Evaluation Metrics

To evaluate the segmentation accuracy quantitatively, we report five metrics: (i) Jaccard index, (ii) Dice coefficient, (iii) average surface distance (ASD), (iv) relative volume error (RVE), and (v) Hausdorff distance (HD) on the ‘frozen’ test mentioned previously. The Jaccard Index measures the intersection over union of the prediction and the GT masks. It is known to be more sensitive to dissimilarity by false positives/negatives more harshly. While the Dice coefficient quantifies the overall agreement between input masks, Jaccard index quantifies what fraction of the prediction correctly overlaps with the GT. HD quantifies the worst-case boundary error (i.e. the largest of all the closest-point distances) unlike ASD, which quantifies the average of all closest-point distances between two surfaces. For clinically oriented assessment of the models, we also computed CSA averaged over C2-C3 vertebral levels of the cervical spinal cord on the predictions from the frozen test set defined previously to measure the morphometric variability for each model. The dataset characteristics of this test set including participant demographics, scanner manufacturers and field strengths are shown in **Table S2**. These measurements are as follows:

1. *CSA*: The per-slice area ($mm^2$) of the predicted segmentation was computed across all slices and then averaged for each contrast.
2. *CSA STD*: For a given contrast, we computed the mean CSA over all slices averaged across the C2-C3 vertebral level. This was repeated for all contrasts for a given participant. Then, across all the participants, we computed the standard deviation (STD) of CSA across all contrasts to assess CSA variability.

The underlying assumption is that one participant should have similar spinal cord CSA across contrasts, with a lower CSA STD corresponding with a better model.

### 2.4.2. Evaluation of segmentations on various contrasts and pathologies

We compared the segmentations between our model's current and previous versions to evaluate the quality of segmentations on challenging cases, including severely compressed spinal cords of DCM patients, and chronic hyperintense lesions of patients with SCI. We also evaluated our model’s ability to produce segmentations on MPRAGE T1map, resting state axial gradient-echo echo-planar-imaging (GRE-EPI) on healthy participants and patients with cervical radiculopathy, whole-spine scans of healthy participants (Molinier et al., 2024) and scans acquired at 7T to highlight the model’s ability to generalize to various MRI contrasts, fields-of-view, scanner strengths, and pathologies unseen during training.

We quantitatively compared the proposed model (contrast_agnostic_v3.0) with its predecessor (contrast_agnostic_v2.0) and existing open-source pathology-specific models `sct_deepseg_sc` (Gros et al., 2019) (for MS) and `SCIsegV2` (Karthik et al., 2024; Naga Karthik, Valošek, et al., 2025) (for SCI and DCM) using the Jaccard Index, Dice Coefficient, Relative Volume Error (RVE), Surface Distance, and Hausdorff Distance from the ANIMA toolbox (Commowick et al., 2018). These models were evaluated on a held-out test set strictly unseen during training following identical preprocessing. The dataset characteristics of pathology specific datasets in Table 1.

Table 1 *Characteristics of the test set of spinal cord pathologies used to evaluate segmentation models. Note the heterogeneity in scanner manufacturers, field strengths and image resolutions. MS=Multiple Sclerosis, SCI=Spinal Cord Injury, DCM=Degenerative Cervical Myelopathy*

| Variables | MS | SCI | DCM |
|---|---|---|---|
| **Number of participants** | 36 | 60 | 30 |
| **Sequences** | T2*w | T2w | T2w |
| **Number of MRI scans** | 36 | 60 | 39 |
| **Sex (Male / Female)** | 16 / 20 | 46 / 13[‡] | 14 / 9[†] |
| **Age (y)** **(mean ± standard deviation)** | N/A | 53.0 ± 15.7 | 59.5 ± 11.5 |
| **Age range (y)** | N/A | 15 – 81 | 37 – 81 |
| **MRI manufacturers** | Philips (n=36) | GE (n=37), Siemens (n=21), N/A (n=2) | Siemens (n=36), N/A (n=3) |
| **MRI field strength** | 3T (n=36) | 1.5T (n=49), 3T (n=9), N/A (n=2) | 3T (n=36), N/A (n=3) |
| **MRI sequence parameters** | in-plane: [0.29 – 0.47] x [0.29 – 0.47] $mm^2$ slice thickness: 2.5 – 5.0 mm | in-plane: [0.35 – 0.8] x [0.35 – 0.8] $mm^2$ slice thickness: 2.75 – 6.0 mm | in-plane: [0.28 – 0.8] x [0.28 – 0.8] $mm^2$ slice thickness: 0.8 – 3.6 mm |

[†] Age/Sex not reported for 7 DCM participants
[‡] Age/Sex not reported for 1 SCI participant

### 2.4.3. Quantitative evaluation of morphometric drift

We applied the proposed lifelong learning framework and quantified the drift in the morphometric variability in terms of the STD of CSA across six contrasts (T2w, T1w, T2*w, MT-on, GRE-T1w, and DWI). Specifically, once released, we let the GitHub Actions workflow run the morphometric analysis and compare our proposed model against two existing spinal cord segmentation methods; `sct_deepseg_sc` (Gros et al., 2019), and `contrast_agnostic_v2.0` (Bédard et al., 2025).

### 2.4.4. Ablation study with recursively-generated GT spinal cord masks

As described in Section 2.2, the spinal cord masks used as GT during training are gathered from multiple sites, containing a combination of manually annotated masks, masks obtained from automatic pathology-specific models. As a result, the differences in delineating the spinal cord-CSF boundary might vary across individual expert raters and the automatic methods due to partial volume effects, hindering model performance. To eliminate this potential noise in the distribution of GT masks gathered from multiple sites, we performed an ablation study where the proposed model was used to produce new GT masks for the entire training set. In practice, this was achieved by running the inference on the entire training dataset and using the automatically generated predictions as the new GT masks for training the subsequent model without any manual corrections. As inter-rater biases are eliminated, the new set of GT masks represents a uniform distribution of GT labels.

### 2.4.5. Updating the normative database of spinal cord morphometrics

The database of healthy adult morphometrics proposed by Valošek et al. (2024) included morphometrics measures computed from 203 healthy individuals from the open-access Spine Generic Multi-Subject dataset (Cohen-Adad et al., 2021b). These morphometric measures were obtained from segmentations generated with `sct_deepseg_sc` (Gros et al. (2019)), with manual corrections for over/under segmentation errors. As outlined in the *Introduction*, morphometric measures are dependent on the segmentation method used. Therefore, we evaluated the following strategy of monitoring and updating the normative database:

1. Generate new segmentations using the proposed `contrast_agnostic_v3.0` model on the T2w scans from 203 healthy participants in the normative database (Valošek et al., 2024).

2. Perform a manual quality control of the spinal cord segmentation masks.

3. Compute 6 morphometric measures(CSA, anteroposterior diameter, transverse diameter, compression ratio, eccentricity and solidity) from the segmentation masks (Valošek et al., 2024).

4. Compute a scaling factor between the morphometric measures derived from different segmentation models, allowing for comparison of morphometric measures across segmentation models.

$$Scaling\ Factor = \frac{metric_{contrast-agnostic_v3.0}}{metric_{sct_deepseg_sc}}$$

#### 2.4.6. Statistical analysis

Statistical analysis was performed using the SciPy (Virtanen et al., 2020), version 1.9.1 and `scikit-posthocs` (Terpilowski, 2019), version 0.10.0. Data normality was tested using the D'Agostino and Pearson's normality test. Between-group comparisons of CSA variability between the proposed and previous models were performed using the non-parametric Friedman test (following significant results from the normality test). Post-hoc tests for pairwise between-group comparisons were performed using the Nemenyi test (following significant results from the Friedman test). To compare the significance of CSA variability between `contrast_agnostic_v3.0` and the model trained on recursively-generated GT masks, we used the non-parametric pairwise Wilcoxon signed-rank test. Unless specified otherwise, $P < .001$ was considered to indicate a statistically significant difference.

# 3. RESULTS

## 3.1. Evaluation on various contrasts and pathologies

#### 3.1.1. Qualitative comparison of segmentations

Figure 4 qualitatively compares the segmentations of `contrast_agnostic_v3.0` (current version), `contrast_agnostic_v2.0` (previous version) and `sct_deepseg_sc` on healthy and pathological scans. While all three models were trained on T1w, T2w and T2*w contrasts, `contrast_agnostic_v2.0` was trained on healthy participant data only and `sct_deepseg_sc` was trained on a multisite dataset of MS patients. We observed a noticeable improvement in the segmentation of the heavily compressed spinal cord (with and without the presence of lesions) in DCM patients with our current model (`contrast_agnostic_v3.0`). Note that `contrast_agnostic_v1.0` is not a model but only a preliminary collection of scripts used to generate the soft ground truths (Bédard et al., 2025).

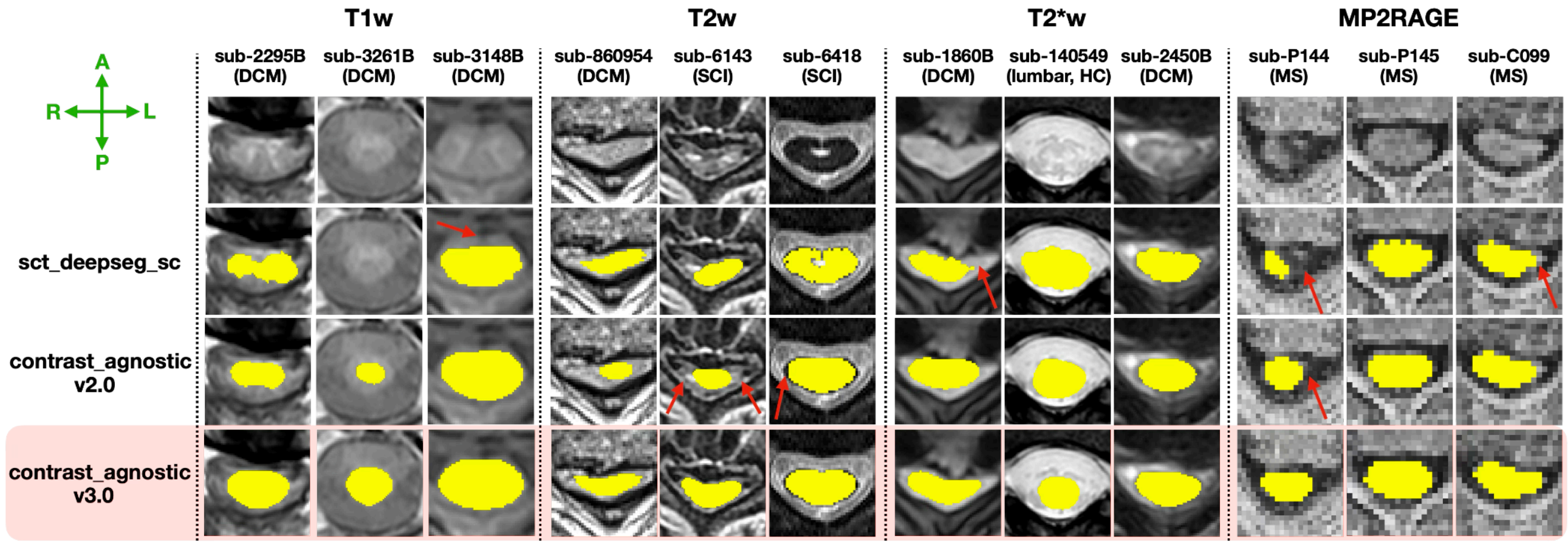

Figure 4 *Comparison of the automatic spinal cord segmentations between* `contrast_agnostic_v3.0` *(current version, highlighted),* `contrast_agnostic_v2.0` *(previous version) and* `sct_deepseg_sc` *on healthy controls (HC), DCM, SCI and MS patients on the test set (unseen during training). Red arrows show the instances where the previous models fail, particularly under heavy compression (with/without lesions) in sub-860594, sub-6143 and sub-1860B.*

Figure 5 qualitatively shows the segmentation outputs of the model across a wide variety of contrasts and pathologies on both sagittal and axial orientations, including whole-spine scans. The model accurately segments the spinal cord under compression (DCM), in cases where the tubular structure of the cord is severely damaged (acute and chronic traumatic SCI) and in the presence of lesions (MS) and atrophy (ALS). All the images used for visualization belong to the test set gathered from different sites (as denoted by different participant IDs in the bottom left) and have never been encountered during training. Notably, in the case of whole-spine images, the model learned to segment the entire spine despite only being trained independently on individual cervical, thoracic and lumbar segments.

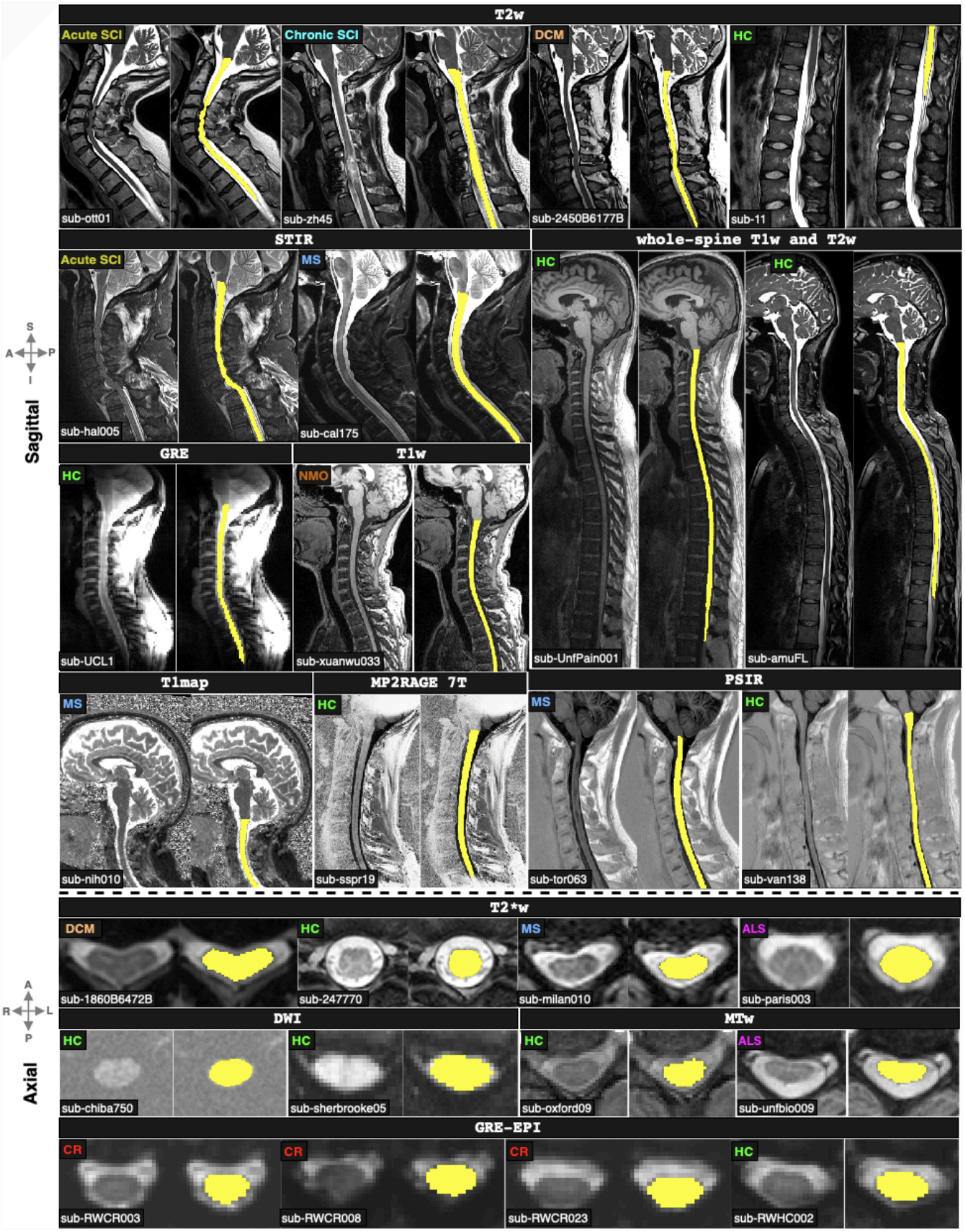

T2w
Acute SCI
sub-ott01
Chronic SCI
sub-zh45
DCM
sub-2450B6177B
HC
sub-11
STIR
Acute SCI
sub-hal005
MS
sub-cal175
whole-spine T1w and T2w
HC
sub-UnfPain001
HC
sub-amuFL
Sagittal
S
A
P
I
GRE
HC
sub-UCL1
T1w
NMO
sub-xuanwu033
T1map
MS
sub-nih010
MP2RAGE 7T
HC
sub-sspr19
PSIR
MS
sub-tor063
HC
sub-van138
T2*w
DCM
sub-1860B6472B
HC
sub-247770
MS
sub-milan010
ALS
sub-paris003
Axial
A
R
L
P
DWI
HC
sub-chiba750
HC
sub-sherbrooke05
MTw
HC
sub-oxford09
ALS
sub-unfbio009
GRE-EPI
CR
sub-RWCR003
CR
sub-RWCR008
CR
sub-RWCR023
HC
sub-RWHC002

Figure 5 *Qualitative visualization of the proposed `contrast-agnostic_v3.0` model's segmentations across various contrasts and pathologies on test images from multiple sites. Our model accurately segments compressed spinal cords, severely damaged cords due to injury, and cords with the presence of lesions. Legend: SCI=spinal cord injury, DCM=degenerative cervical myelopathy, MS=multiple sclerosis, NMO=neuromyelitis optica, ALS=amyotrophic lateral sclerosis, CR=cervical radiculopathy, and HC=healthy control.*

### 3.1.2. Quantitative evaluation on healthy controls and pathologies

Table 2 presents a quantitative comparison of the current (contrast_agnostic_v3.0) and previous (contrast_agnostic_v2.0) versions of the segmentation model in the lifelong training framework along with the existing pathology-specific models on test sets gathered from multiple sites containing healthy participant and pathological data. Starting with a comparison of the models on the frozen test set of healthy participants (Table 2**A**), we then present results for test sets containing T2w and T2*w images of the lumbar cord of healthy participants from two sites (Table 2**B**), T2*w images of MS patients from two sites (Table 2**C**), axial and sagittal T2w scans of DCM patients from two sites (Table 2**D**), and axial and sagittal T2w scans of traumatic SCI (acute, intermediate and chronic phases) from six sites (Table 2**E**). In all comparisons, the proposed `contrast_agnostic_v3.0` model achieved similar or better performance compared to the previous state-of-the-art or pathology-specific models with major improvement in the segmentation of the lumbar cord.

Table 2 *Quantitative comparison of spinal cord segmentations for previous segmentation methods on test sets containing: (a) healthy participants (cervical cord) (n = 49 participants; $n_{vol}$ = 294 images) averaged across all contrasts, (b) lumbar scans of healthy participants (n=9), (c) patients with MS on T2*w contrast (n = 36 participants; $n_{vol}$ = 36 images), (d) DCM on axial and sagittal T2w scans ($n_{vol}$ = 39), and (e) SCI on axial and sagittal T2w scans ($n_{vol}$ = 60). RVE stands for Relative Volume Error, ASD stands for Average Surface Distance, and HD stands for Hausdorff Distance.* ***Bold*** *values represent the best-performing model for that metric.*

| **Methods** | **Jaccard (↑)** | **Dice (↑)** | **RVE (%)** | **ASD (↓)** | **HD (↓)** |
|---|---|---|---|---|---|
| | Opt. value: 1 | Opt. value: 1 | Opt. value: 0 | Opt. value: 0 | Opt. value: 0 |
| **A) Healthy participants (cervical cord) (n = 49; 6 contrasts per participant; $n_{vol}$ = 294)** | | | | | |
| sct_deepseg_sc | 0.91 ± 0.05 | 0.95 ± 0.03 | -0.18 ± 8.95 | 0.04 ± 0.27 | 3.18 ± 6.16 |
| contrast_agnostic _v2.0 | 0.91 ± 0.03 | 0.95 ± 0.02 | **-0.05 ± 4.18** | **0.02 ± 0.12** | **2.68 ± 4.12** |
| contrast_agnostic | **0.92 ± 0.03** | **0.96 ± 0.02** | -0.76 ± 4.59 | 0.04 ± 0.27 | 3.02 ± 6.41 |
| _v3.0 (proposed) | | | | | |
| **B) Healthy participants (lumbar cord) (n = 9; T2w and T2*w; $n_{vol}$ = 9)** | | | | | |
| sct_deepseg_sc | 0.67 ± 0.13 | 0.78 ± 0.11 | 57.56 ± 53.28 | 0.34 ± 0.58 | 61.87 ± 101.46 |
| contrast_agnostic _v2.0 | 0.60 ± 0.24 | 0.71 ± 0.26 | 32.47 ± 32.94 | 7.14 ± 18.58 | 57.04 ± 84.45 |
| contrast_agnostic _v3.0 (proposed) | **0.86 ± 0.02** | **0.93 ± 0.01** | **-3.61 ± 4.78** | **0.015 ± 0.02** | **7.48 ± 7.42** |
| **C) Patients with MS (n = 36; T2*w contrast; $n_{vol}$ = 36)** | | | | | |
| sct_deepseg_sc | 0.89 ± 0.03 | 0.94 ± 0.02 | -9.03 ± 3.35 | **0.003 ± 0.009** | 3.06 ± 1.17 |
| contrast_agnostic _v2.0 | 0.88 ± 0.03 | 0.94 ± 0.01 | -10.12 ± 2.89 | 0.009 ± 0.016 | 3.43 ± 1.61 |
| contrast_agnostic _v3.0 (proposed) | **0.92 ± 0.03** | **0.96 ± 0.01** | **-5.34 ± 2.89** | 0.005 ± 0.014 | **2.96 ± 2.39** |
| **D) Patients with DCM (n = 32; T2w contrast; $n_{vol}$ = 39)** | | | | | |
| SCIsegV2 | **0.94 ± 0.02** | **0.97 ± 0.01** | **-2.34 ± 1.79** | 0.001 ± 0.001 | 3.13 ± 1.67 |
| contrast_agnostic _v2.0 | 0.84 ± 0.04 | 0.91 ± 0.02 | -11.91 ± 4.16 | 0.01 ± 0.04 | 6.23 ± 4.83 |
| contrast_agnostic _v3.0 (proposed) | 0.93 ± 0.02 | 0.96 ± 0.01 | -2.51 ± 2.25 | **0.001 ± 0.001** | **3.14 ± 1.62** |
| **E) Patients with SCI (n = 60; T2w contrast; $n_{vol}$ = 60)** | | | | | |
| SCIsegV2 | **0.86 ± 0.06** | **0.93 ± 0.04** | 5.22 ± 7.63 | **0.01 ± 0.01** | 12.83 ± 23.51 |
| sct_deepseg_sc | 0.73 ± 0.23 | 0.82 ± 0.23 | -13.68 ± 24.1 | 7.61 ± 31.87 | 38.39 ± 77.35 |
| contrast_agnostic _v2.0 | 0.60 ± 0.17 | 0.74 ± 0.17 | -28.81 ± 20.49 | 1.38 ± 4.56 | 64.27 ± 85.62 |
| contrast_agnostic _v3.0 (proposed) | 0.87 ± 0.08 | **0.93 ± 0.06** | **1.75 ± 14.63** | 0.01 ± 0.04 | **10.3 ± 22.6** |

## 3.2. Quantitative evaluation of morphometric drift across model versions

### 3.2.1. Variability of CSA across contrasts

The figures below are automatically output by the GitHub Actions workflow in the proposed lifelong training framework.

Figure 6 shows the CSA STD across six contrasts on the test set of healthy participants (*n=49*; $n_{vol}$*=294*) of the spine-generic Multi-Subject database (Cohen-Adad et al., 2021b) between three methods: (i) `sct_deepseg_sc` (Gros et al., 2019), (ii) our previous version, `contrast-agnostic_v2.0` (Bédard et al., 2025), and the current version, `contrast-agnostic_v3.0`. The `contrast-agnostic_v3.0` model obtained relatively more stable segmentations with the lowest STD of CSA across contrasts compared to the other methods. **Figure S1** plots the variability in spinal cord CSA per each individual contrast. Similar to the analysis of CSA variability across contrasts, we also plot the variability in CSA across 3 vendors (GE, Siemens, and Philips) on a test set containing scans of a healthy participant acquired from 15 sites in **Figure S2**.

In Figure 7, we plot the level of agreement between the CSA estimated by the models on the commonly used T1w and T2w contrasts on the same test set described above. In addition to segmenting a wide range of contrasts and pathologies as shown in the previous figures, the `contrast-agnostic_v3.0` model achieves a similar alignment between T1w and T2w contrasts as our previous model trained only on a healthy participant database. Table 2**A** compares the models' performances using quantitative metrics.

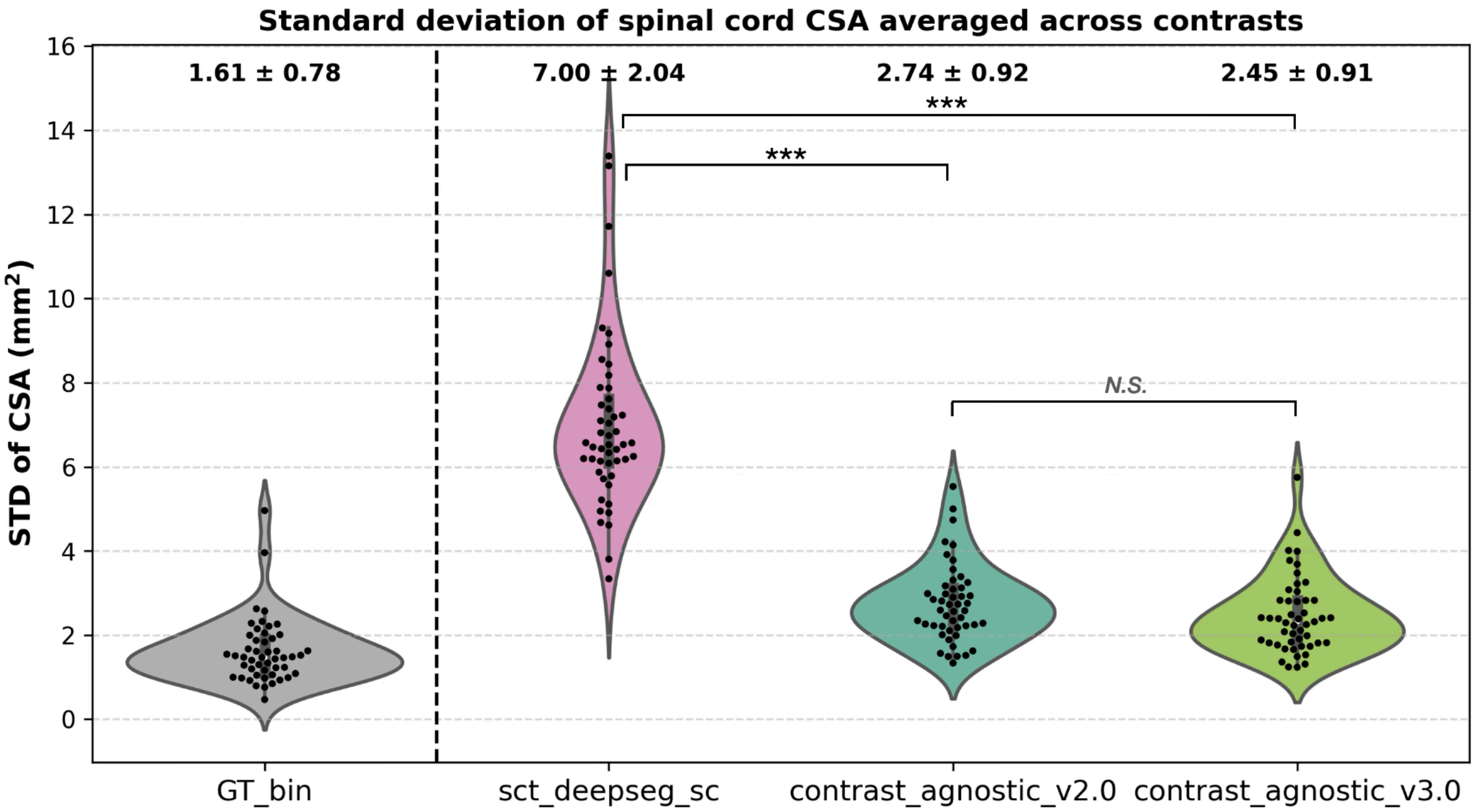

Figure 6 *CSA variability measured in terms of the standard deviation across 6 contrasts on a test set of healthy participants (n=49). The lower the CSA STD across contrasts, the better. Our proposed model achieved the lowest STD averaged across 6 contrasts (i.e. each point shows the mean of 6 contrasts for a given participant) showing more stability in segmentations across contrasts. *** P < 0.001 (non-parametric Friedman test followed by post-hoc Nemenyi test for pairwise comparisons). "N.S.": Non significant. Note that there is no statistically significant difference between `contrast_agnostic_v2.0` and `contrast_agnostic_v3.0`, implying that CSA variability achieved by the proposed model is at least as good as its previous version despite the addition of a diverse set of contrasts and pathologies.*

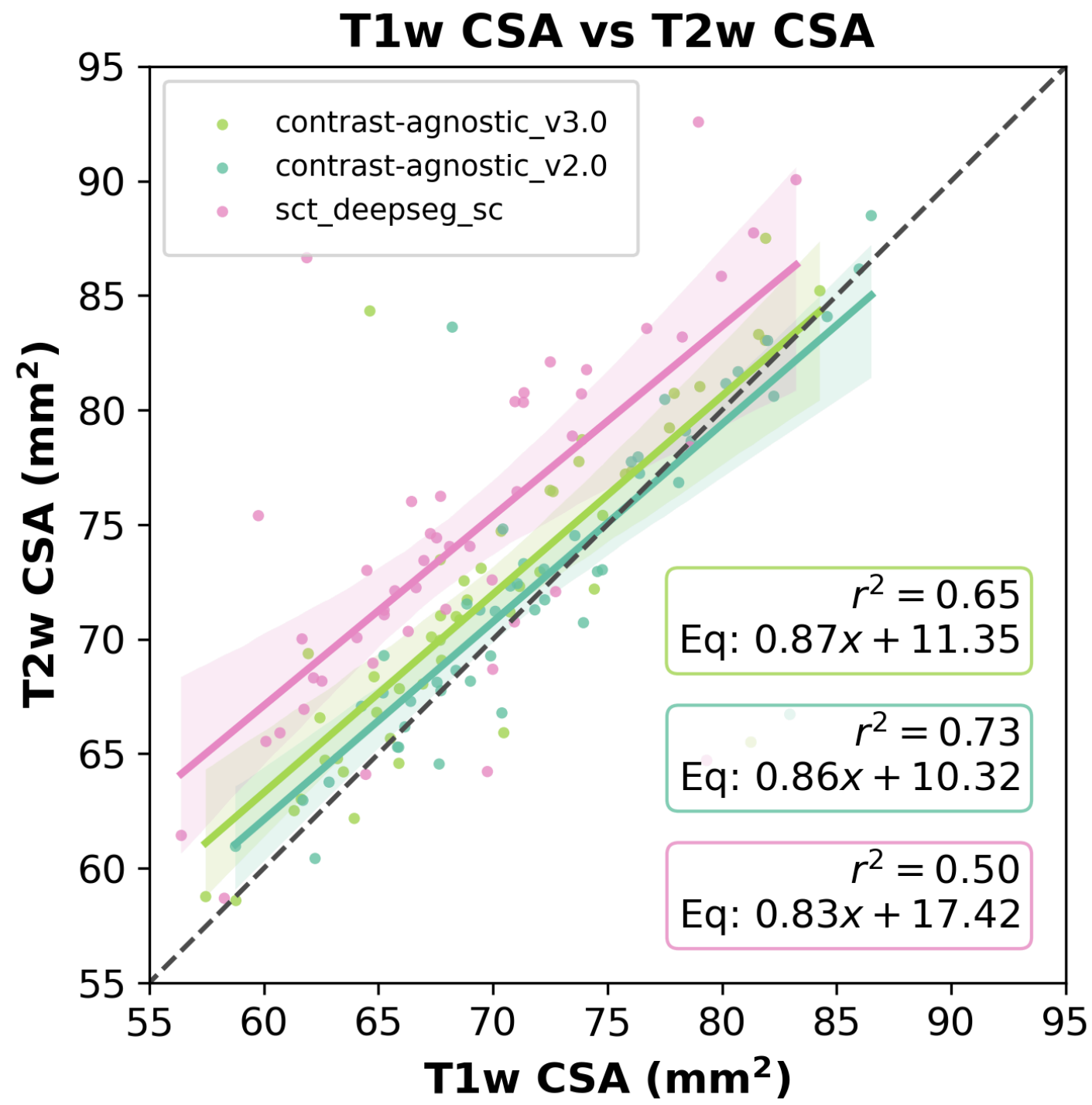

Figure 7 *Level of agreement between T1w and T2w CSA at C2-C3 for `contrast_agnostic_v3.0`, `contrast_agnostic_v2.0` and `sct_deepseg_sc`. Each point represents one test participant (n=49). The dashed diagonal line in black represents perfect agreement between the CSA of T1w and T2w contrasts. Points in the upper left triangle of the diagonal line represent overestimation of T2w CSA. Note the sct_deepseg_sc heavily overestimates T2w CSA for majority of test participants, whereas contrast agnostic models v2.0 and v3.0 lower the CSA variability.*

### 3.2.2. CSA variability with recursively-generated GT masks

Since the GT masks for each contrast and pathology in the training set are a mixture of manual segmentations from different raters and automatic segmentations from different models, the collection of GT masks can be seen as a noisy distribution of segmentations with high variability at the spinal cord-CSF boundary. Figure 8 shows the results of our ablation study where all the GT masks were re-generated with `contrast-agnostic_v3.0`, and a new model was trained on the resulting collection. Recall that no manual corrections (or QC) were performed to maintain a uniform distribution of the regenerated GT masks. We used the same test set of healthy participants (*n=49, 6 contrasts*) from the spine-generic multi-subject database and compared two models: (i) the proposed model with the original (noisy) distribution of GT masks (shown with the green violin plot), and (ii) the proposed model, but trained on the new (uniform) distribution of the GT masks (shown with the blue violin plot). We observed that the model trained on the recursively generated GT masks showed a slightly higher STD across contrasts compared to the model trained on the original GT masks. In supplementary **Figure S3**, we also plot the variability in CSA per contrast between the two methods, demonstrating how the model trained on recursively generated GT masks underestimated the CSA on all contrasts.

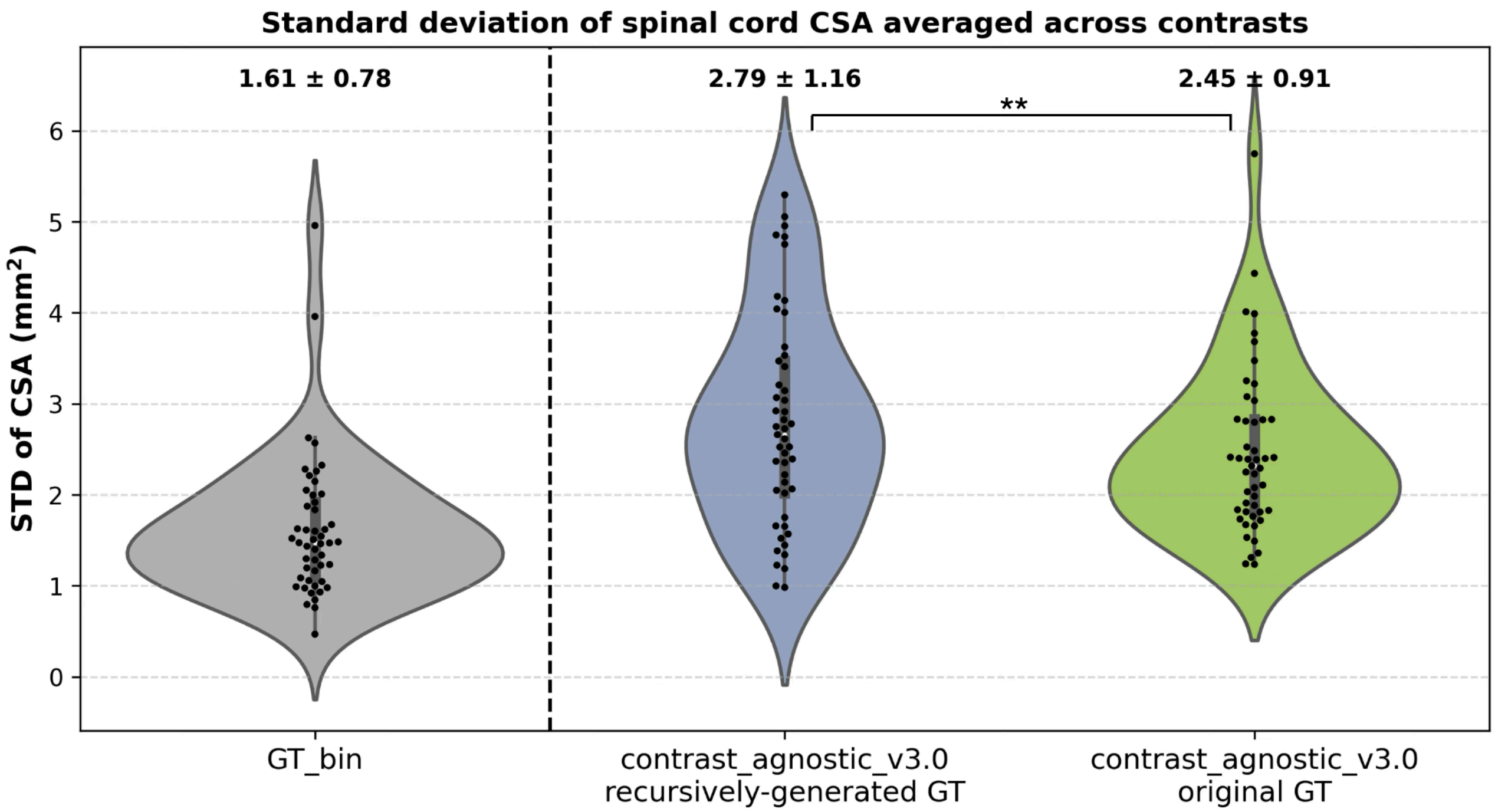

Figure 8 *Standard deviation of the CSA across 6 contrasts for models trained on: (i) recursively generated GT masks (blue), and (ii) original GT masks (green). Each point shows the mean of 6 contrasts for a given participant. The lower the CSA STD across contrasts, the better. Note that the model trained on original GT masks (combination of manual and automatic segmentations) tends to produce stable segmentations resulting in a lower STD across contrasts compared to the uniform distribution of recursively generated GT masks. ** P < 0.01 (two-sided non-parametric Wilcoxon signed-rank test).*

### 3.2.3. Normative database results

Figure 9**A** shows the plots for 6 different morphometric measures computed on 203 healthy participants using two versions of segmentation masks: (i) the segmentations from `sct_deepseg_sc` with manual corrections (pink) used in (Valošek et al., 2024) and (ii) the segmentations from the proposed `contrast-agnostic_v3.0` model (green, no manual correction). Given the difference in the segmentations at the cord-CSF boundary, we present the *scaling factor* between the morphometric measures computed with the 2 methods in Figure 9**B**.

We observed that the scaling factor is nearly constant among slices across the given vertebral levels. For the benefit of future studies using the normative database of spinal cord morphometrics, they have been made open-source[4].

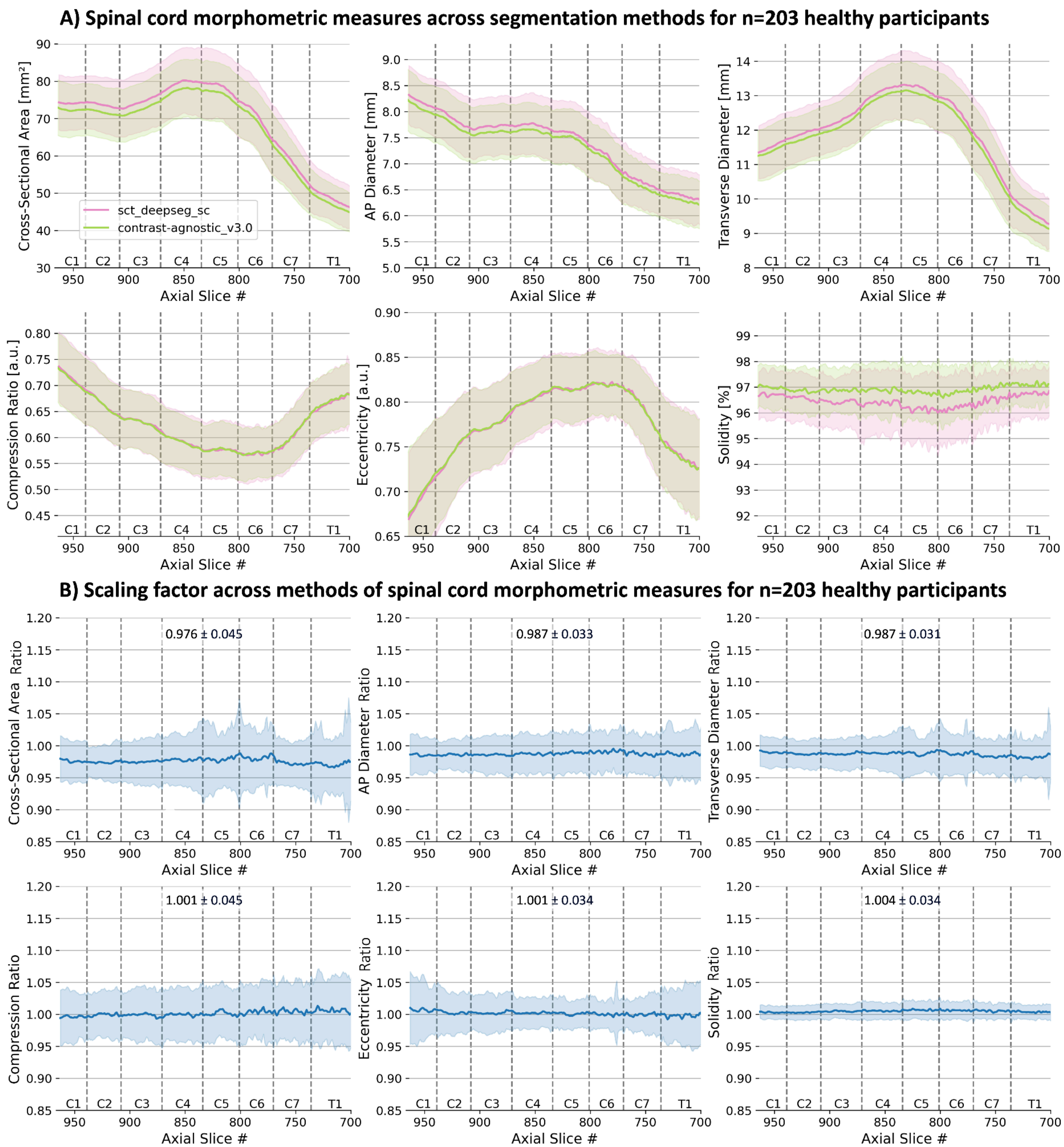

[4] https://github.com/spinalcordtoolbox/PAM50-normalized-metrics/releases/tag/r20250321

Figure 9 ***(A)*** *Morphometric measures computed on n=203 healthy participants from the Spine Generic Dataset (Cohen-Adad et al., 2021b) for 6 morphometric measures using 2 different segmentation methods:* `sct_deepseg_sc` *with manual correction (green) and* `contrast-agnostic_v3.0` *(pink) with* ***(B)*** *scaling factor between the methods (means ± std). Metrics are shown in the PAM50 space. Note that despite the diversity of the training dataset containing various contrasts and pathologies, the proposed v3.0 model achieved nearly constant scaling factors for all morphometric measures compared to the v2.0 model which was trained only on a healthy participant database.*

# 4. DISCUSSION

In this study, we presented an automatic model for the robust segmentation of the spinal cord across different MRI contrasts and pathologies. Our model was developed using heterogeneous data gathered from 75 clinical sites and hospitals worldwide, acquired with different resolutions, orientations, field strengths, and scanner manufacturers. We have shown that our proposed model provides reliable spinal cord segmentation on MRI scans across different pathologies including spinal cord compression (asymptomatic compression and DCM), atrophy (ALS), severely injured spinal cords in traumatic SCI, and spinal cords containing intramedullary lesions (SCI and MS). To facilitate the continual development of segmentation models over time, we presented a *lifelong learning* scenario to automatically monitor the drift in morphometric variability across various model versions and enable periodic retraining by adding new contrasts and pathologies. As a real-world application of the lifelong learning framework, we applied the most recent version of our spinal cord segmentation model to update the morphometric measures of a normative database of healthy adults.

## 4.1. Data curation

Data gathered from multiple sites tend to be noisy in many respects, due in part to various imaging artifacts, metallic hardware, and environmental noise. While noisy GT masks are inevitable due to inter-rater variability, they could potentially be useful for training robust segmentation models (Shi & Wu, 2021; Yao et al., 2023). However, noise in training data tends to disrupt model training by making the models unintentionally focus on such outliers (Rahman et al., 2022; Taha & Hanbury, 2015), resulting in poor overall segmentation and inaccurate evaluation of the models' performance. In our proposed lifelong training scenario, it was critical to ensure the quality of the input data at each step of model development over time, as our segmentation models were trained from scratch on all the previous and new data. To account for this, we labelled each new dataset containing new contrasts or pathologies with existing automatic models (Bédard et al., 2025; Gros et al., 2019; Naga Karthik, Valošek, et al., 2025) and used

sct_qc (SCT's visual QC tool) to quickly identify cases with failed segmentations requiring manual corrections and flagged images with strong artifacts for exclusion. These QC reports provide compressed snapshot of the datasets, which is useful for sharing with the clinical sites (Jwa et al., 2025).

Inter-rater variability was not quantitatively assessed in this study. Most manual segmentations were done by single raters at their respective sites with no travelling/overlapping participants/raters across sites. However, evidence from the literature shows that automated methods perform on par with the inter-rater discrepancy. De Leener et al. (De Leener et al., 2014) found PropSeg predictions and individual manual segmentations from 2 raters had similar Dice scores, comparable to inter-rater variability. In a clinical study, Connor et al. (Connor et al., 2025) found 90.8% agreement between manual and automated measures of lesion damage following SCI on T2-w scans. In a similar set-up, Veiga-Canuto et al. (Veiga-Canuto et al., 2022) compared manual and nnUNet-based automatic segmentations of neuroblastic tumors in MRI. They found the automatic method achieved similar inter-rater variability to manual raters, with a median false-negative ratio of 0.96. Given the fact that the spinal cord is a large, easily detectable object compared to smaller objects such as lesions or tumors, we anticipate high agreement between the proposed contrast_agnostic_v3.0 and inter-rater variability, consistent with prior research.

## 4.2. Lifelong learning segmentation of the spinal cord

### 4.2.1. Robustness across contrasts and pathologies

Gathering datasets containing new contrasts and pathologies over time, and training a model on this aggregated dataset resulted in robust segmentation of the spinal cord on a wide range of contrasts and pathologies. As seen in Figure 4 and Figure 5, the contrast_agnostic_v3.0 model performed comparatively well when measured against the performance of previous models when applied to unseen images, benefitting from the lifelong learning strategy of updating the training database with new contrasts and pathologies. This was particularly notable for samples which exhibited severe compression, in the presence of both hyper/hypo-intense lesions (MS and its phenotypes and different SCI phases), on lumbar spine, and unusual scanner strengths (7T MP2RAGE). Our model also performed well by generalizing to MRI contrasts not included in the training set (e.g., MPRAGE T1map, GRE-EPI, and Fieldmap images). Interestingly, the model was also capable of accurate whole-spine segmentation, despite only being trained on "chunks" of individual spinal regions. This echoes the findings of our recent study, which found that segmentation models do not benefit from additional context when trained on scans covering the entire spinal cord (Naga Karthik, McGinnis, et al., 2025).

The competitive performance of the proposed model compared to existing pathology-specific models (Table 2) highlights the advantage of continually developing segmentation models over time as it reduces the cost of maintaining multiple models while ensuring that single class of models can be trained to be contrast- and pathology-agnostic over time.

### 4.2.2. Automatic monitoring of morphometric drift

Continuous monitoring of deployed models in production is a standard practice in MLOps pipelines, achieved through software technologies such as Docker, GitHub Actions, Kubernetes, and Git LFS (Kandpal et al., 2023; Spjuth et al., 2021; Tabassam, 2023). In a continuous learning system, monitoring deployed models is critical to ensure that the performance of the models on downstream tasks does not significantly degrade throughout their evolution (Agirre et al., 2021). Performance drifts could be caused by shifts in the input data distribution, typically manifesting in the form of changes in the participant demographics (e.g. adult population to pediatric population) and acquisition parameters (e.g. 3T data to 7T data) (González et al., 2024). Therefore, monitoring morphometric drift between various model versions is crucial, as downstream tasks which rely on quantifying changes in the spinal cord morphometry are strongly tied to the accuracy of the segmentation (Joo et al., 2025; Valošek et al., 2024). In this regard, our proposed automatic workflow for monitoring morphometric drift provides a quick feedback loop with two possible outcomes: (i) the magnitude of drift in the CSA variability with the new model is high, thus requiring re-evaluation of the data curation and/or model training steps to bring the drift within an acceptable range, or, (ii) the magnitude of CSA drift is within an acceptable range of the previous "stable" version, making it the new state-of-the-art for annotating (new) unlabelled data to train subsequent models. In the first outcome, we note that the CSA drift threshold was not defined apriori and that model retraining criteria was judged using CSA STD plots (Figure 6) and qualitative evaluation of model's predictions at the cord boundaries with QC reports. As the test set required to be publicly-accessible for the GitHub workflow *and* only contain the images of healthy participants from six contrasts to ensure fair comparison with previous model, we observed that additional QC in challenging pathological cases such as the ones with lesions and/or heavily-compressed cords was critical to ensure that the model learned to segment the spinal cord boundaries appropriately.

Another important note is that the morphometric drift was primarily assessed by measuring the CSA across contrasts. While antero-posterior (AP) and right–left (RL) diameters could also capture changes related to boundary positioning, they would essentially reflect similar underlying information as CSA. Other shape-based metrics, such as eccentricity or compression ratio measure the overall shape of the cord (i.e. its ellipticity) or ratios of AP/RL diameters but are not suitable for measuring systematic shifts in cord boundaries.

Also, note that the proposed lifelong learning framework using GitHub Actions is not specific to spinal cord segmentation but can be reused for any other segmentation task involving the development of multiple models over time.

### 4.2.3. Training on recursively generated labels

Any form of human intervention is undesirable in a post-deployment lifelong learning scenario making it prone to errors. However, existing models are unable to automatically utilize incoming data as it arrives (Agirre et al., 2021; González et al., 2024; Prapas et al., 2021), necessitating periodic checks to prevent degradation of model performance. While our proposed continuous training strategy automatically monitors the drift in morphometric variability *after* training, one could also automate the re-training process, thus making the continuous learning loop *fully automatic.* Currently, when new data arrives, we rely on the combination of automatic annotation using the latest version of the model and performing visual QC, identifying cases with failed/incorrect segmentations for manual corrections. What if we forego this data curation step involving manual intervention?

In our attempt to evaluate the potential of such an approach (Figure 8, **Figure S3**), we observed that the model underestimated the average CSA on a healthy subset of participants for each of the 6 contrasts and resulted in a significantly higher CSA STD across contrasts ($P < 0.01$), when compared to the performance of the model trained on the original GT masks obtained from a combination of automatic and manual segmentations. Recent research in the context of text generation and image synthesis (Shumailov et al., 2024) has shown that multiple iterations of training on recursively generated data tend to make the model *catastrophically forget* (Sodhani et al., 2022) the underlying true data distribution, leading to model collapse, something which we did not observe. Given the inconsistencies in manual/automatic segmentations at the cord-CSF boundary owing to varying partial volume effects with images of different contrasts and resolutions, we hypothesize that training on such noisy labels acted as an inherent regularizer, making the model more robust across contrasts. On the other hand, training on uniform distribution of model-generated segmentations where the inconsistencies have been smoothed out, the model tends to under-segment the spinal cord, something which would need to be kept in mind for analyses based on models trained this way.

Successive training with recursively-generated GT masks can potentially propagate errors in the training data distribution. To quantitatively assess bias introduced by recursive use of automatic segmentations, an initial baseline (X%) is established by evaluating the original model ($M_0$) on a manually annotated test set ($T_0$). A subsequent model ($M_1$) is then trained using masks automatically generated by $M_0$ and evaluated on (i) the auto-generated masks (Y%) and (ii) the original manual set ($T_0$) (Z%). The discrepancy (Y–Z)% quantifies the degree of bias, with larger values indicating that $M_1$ has overfit to systematic errors in the recursive ground truth. This experiment is difficult to scale, as fully manual annotations remain time-consuming and costly to produce.

#### 4.2.4. Binary vs. soft masks

While training directly on soft masks still achieves the lowest morphometric variability across contrasts (Bédard et al., 2025), the registration step (which requires mutual co-registration of all contrasts) requires more than one contrast per participant, becoming a bottleneck in developing segmentation models, as well as further manual intervention in correcting registration outputs across both healthy and pathological data. Furthermore, training on *soft* masks requires converting existing datasets with binary GT masks to soft masks within an appropriate contrast-dependent threshold. Given the lifelong learning framework for developing segmentation models, the *softness* of the masks from one model cannot be accurately quantified to match the softness for the next model, owing to partial volume effects and differences in the training data distribution, subtly biasing the ground truth with subsequent newer versions of the model. On the contrary, training on binarized GT masks (thresholded at 0.5) presents a simple and scalable solution, reducing the impact of model-specific biases as most models tend to be uncertain at the boundaries of the segmentation masks (Lemay et al., 2022). While training on binary masks is scalable in a lifelong learning framework, it could potentially be limiting in cases where the CSA is small at the tip of the spinal cord. In these regions, soft masks can better represent the partial volume compared to binary masks. We refer the reader to (Bédard et al., 2025) for detailed qualitative and quantitative comparisons on the effects of training with soft and binary masks and the usage of pixel-wise classification and regression-based loss functions towards the estimation of CSA.

### 4.3. Application on normative database of morphometrics

Keeping an updated normative morphometrics database is crucial to maintaining lifelong models (Valošek et al., 2024), as it allows users to relate their measurements obtained using the latest segmentation method up-to-date. Additionally, when adding new individuals to the normative database, one should re-segment all images within it using the latest segmentation method to ensure the database follows the state of the model. Maintaining and updating such a dataset requires coordination across the segmentation model, the SCT software, and the Spine Generic dataset, a process not currently implemented, but can be accomplished using GitHub Actions. The scaling factors identified using our framework also ensures backward compatibility with previous segmentation methods included in SCT (i.e., `sct_deepseg_sc`), allowing researchers to compare morphometric measures derived from different segmentation models. We encourage users to update to `contrast_agnostic_v3.0`, however, as it significantly improves the spinal cord segmentation robustness in previously difficult pathologies, such as cord compression and spinal cord injury.

### 4.4. Limitations

A major limitation of this study is that our strategy for monitoring and evaluating morphometric drift across various model versions depends on a fixed set of contrasts (*n=6*) in a frozen test set of healthy participants. While newer models may generalize well to other pathologies and contrasts, their true performance could be limited by the evaluation of the CSA on only 6 contrasts. Future work could add better methods for evaluating morphometric drift (e.g. by computing other commonly used spinal cord morphometrics) on data from both healthy participants and from participants with spinal cord pathologies. With the rise of open-source challenges targeting specific spinal cord pathologies[5,6], our GitHub Actions-based workflows could be adapted to include evaluations not only of healthy participants but on participants with pathologies as well.

Stagnation of the training data distribution when developing models over time is another issue. With subsequent models being trained on new data (potentially from different populations – pediatric, adult and geriatric), the data distribution used for the earliest model might no longer be representative of the current distribution. In such cases, comparing histogram-based distribution shifts using KL divergence, or detecting drifts in the feature space by extracting radiomic features (van Griethuysen et al., 2017) could ensure the continued relevance of the training and test sets for evaluating future models. If the drift between data distribution is large, keeping only a subset of the old data when training new models is recommended.

## 5. CONCLUSION

This study introduces an automatic tool for the robust segmentation of the spinal cord across various MRI contrasts and spinal pathologies. The model was trained on diverse datasets collected from 75 clinical sites and hospitals worldwide, with heterogeneous image resolutions, orientations, field strengths, and scanner manufacturers. Our results demonstrate that the model effectively segments spinal cord scans from healthy participants, as well as from those with compressions, atrophy, intramedullary lesions and SCI. To support the continuous improvement of segmentation models, we propose a lifelong learning framework which automatically monitors the drifts in morphometric variability across model versions. The proposed framework facilitates periodic retraining by incorporating new contrasts and pathologies and provides a quick feedback loop for developing future segmentation models. As a real-world application of this framework, we employed the proposed spinal cord segmentation model to update morphometric measurements in a normative database of healthy adults. Our results showed that the scaling factor required to update the database of morphometric measures is nearly constant among slices across the given vertebral levels, showing minimum drift between the current and previous versions of the model trained within the lifelong learning framework.

[5] https://portal.fli-iam.irisa.fr/ms-multi-spine/

[6] https://ivdm3seg.weebly.com

# 6. ACKNOWLEDGEMENTS

We thank Nick Guenther and Mathieu Guay-Paquet for their assistance with the management of the datasets, Joshua Newton for his contributions in helping us implement the algorithm to SCT, and Kalum Ost for reviewing and editing this manuscript. We thank Drs. Thierry Albert, Bertrand Baussart, Caroline Hugeron, Hugues Pascal Moussellard, Frédéric Petit and Marc-Antoine Rousseau for helping with patient recruitment in Paris, Dr. Serge Rossignol and the Multidisciplinary Team on Locomotor Rehabilitation (Regenerative Medicine and Nanomedicine, CIHR), and Drs. Jean Pelletier and Bertrand Audoin, and Lauriane Pini, for data acquisition in Marseille. Lastly, we thank all the participants including the RHSCIR participants and network, from all the following participating local RHSCIR sites: Vancouver General Hospital, Foothills Hospital, Royal University Hospital, Toronto Western Hospital, St. Michael's Hospital, Sunnybrook Health Sciences Centre, Hamilton General Hospital, The Ottawa Hospital Civic Campus, Hôpital de l'Enfant Jésus, Hôpital du Sacre Coeur de Montréal, QEII Health Sciences Centre, and Saint John Regional Hospital. For more information about RHSCIR, please visit www.praxisinstitute.org.

# 8. ETHICS

All participants in their respective imaging sites signed a consent form as per the local IRB regulations. Data were then aggregated at the managing site, under Polytechnique Montréal's IRB (CER-2324-26-D).

# 9. DATA AND CODE AVAILABILITY

The datasets used in this study consist of a mix of publicly-available (✓) and privately-held datasets. The dataset identifiers used internally along with the accessibility are as follows: basel-mp2rage (✗), canproco (✗), data-multi-subject (✓), dcm-brno (✗), dcm-zurich-lesions-20231115 (✗), dcm-zurich-lesions (✗), dcm-zurich (✗), lumbar-epfl (✗), lumbar-vanderbilt (✗), nih-ms-mp2rage (✗), sci-colorado (✗), sci-paris (✗), sci-zurich (✗), sct-testing-large (✗), site_006_praxis (✗), site_007_praxis (✗). Publicly-available datasets are cited using their corresponding publication wherever applicable. The code for this study is open-source and can be accessed at https://github.com/sct-pipeline/contrast-agnostic-softseg-spinalcord/releases/tag/v3.0 (which also contains the overview of dataset characteristics). For ease of accessibility, the spinal cord segmentation model is integrated into Spinal Cord Toolbox (v7.0 and above). The updated database of spinal cord morphometric measures can be accessed at https://github.com/spinalcordtoolbox/PAM50-normalized-metrics/releases/tag/r20250321.

# 10. AUTHOR CONTRIBUTIONS

E.N.K.: data curation, formal analysis, investigation, methodology, visualization, and writing (original draft, review and editing). S.B.: data curation, formal analysis, investigation, methodology, visualization, and writing (original draft, review and editing). J.V.: data curation, formal analysis, investigation, methodology, visualization, and writing (original draft, review and editing). J.C.A.: conceptualization, data curation, funding acquisition, investigation, methodology, supervision, writing (review and editing). The rest of the authors: data curation, writing (review and editing).

# 11. DECLARATION OF COMPETING INTERESTS

Conflicts of interest outside the present work: M. Filippi is Editor-in-Chief of the *Journal of Neurology*, Associate Editor of *Human Brain Mapping*, *Neurological Sciences,* and *Radiology*, received compensation for consulting services from Alexion, Almirall, Biogen, Merck, Novartis, Roche, Sanofi, speaking activities from Bayer, Biogen, Celgene, Chiesi Italia SpA, Eli Lilly,

Genzyme, Janssen, Merck-Serono, Neopharmed Gentili, Novartis, Novo Nordisk, Roche, Sanofi, Takeda, and TEVA, participation in Advisory Boards for Alexion, Biogen, Bristol-Myers Squibb, Merck, Novartis, Roche, Sanofi, Sanofi-Aventis, Sanofi-Genzyme, Takeda, scientific direction of educational events for Biogen, Merck, Roche, Celgene, Bristol-Myers Squibb, Lilly, Novartis, Sanofi-Genzyme, he receives research support from Biogen Idec, Merck-Serono, Novartis, Roche, the Italian Ministry of Health, the Italian Ministry of University and Research, and Fondazione Italiana Sclerosi Multipla.

M.A. Rocca received consulting fees from Biogen, Bristol Myers Squibb, Eli Lilly, Janssen, Roche; and speaker honoraria from AstraZeneca, Biogen, Bristol Myers Squibb, Bromatech, Celgene, Genzyme, Horizon Therapeutics Italy, Merck Serono SpA, Novartis, Roche, Sanofi and Teva. She receives research support from the MS Society of Canada, the Italian Ministry of Health, the Italian Ministry of University and Research, and Fondazione Italiana Sclerosi Multipla. She is Associate Editor for *Multiple Sclerosis and Related Disorders*; and Associate Co-Editor for Europe and Africa for *Multiple Sclerosis Journal*.

E. Pravatà received honoraria for serving on scientific advisory boards from Bayer AG, unrelated to the present work. C. Mainero and C.A. Treaba received research support from Genentech. Daniel S. Reich has received research funding from Abata and Sanofi. The rest of the authors declared no potential conflicts of interest with respect to the research, authorship, and/or publication of this article.

## 12. FUNDING

ENK is supported by the Fonds de Recherche du Québec Nature and Technologie (FRQNT) Doctoral Training Scholarship. SB is supported by the Natural Sciences and Engineering Research Council of Canada, NSERC, Canada Graduate Scholarships — Doctoral program. JV received funding from the European Union's Horizon Europe research and innovation programme under the Marie Skłodowska-Curie grant agreement No 101107932. Funded by the Canada Research Chair in Quantitative Magnetic Resonance Imaging [CRC-2020-00179], the Canadian Institute of Health Research [PJT-190258], the Canada Foundation for Innovation [32454, 34824], the Fonds de Recherche du Québec - Santé [322736, 324636], the Natural Sciences and Engineering Research Council of Canada [RGPIN-2019-07244], the Canada First Research Excellence Fund (IVADO and TransMedTech), the Courtois NeuroMod project, the Quebec BioImaging Network [5886, 35450], INSPIRED (Spinal Research, UK; Wings for Life, Austria; Craig H. Neilsen Foundation, USA), Mila - Tech Transfer Funding Program, the Digital Research Alliance of Canada (alliancecan.ca) for providing the supercomputing infrastructure, NIH R01NS078322 (CM). JV, JB, TH and PK received funding from Czech Health Research Council grants NV18-04-00159, NU22-04-00024 and LM2018129 Czech-BioImaging. DVDV, IR and NK received funding from the Swiss National Science Foundation grant [205321_207493]. DP is funded by a Swiss National Science

Foundation Postdoc Mobility Fellowship grant [P500PM_214211]. CT is founded by the Swiss National Science Foundation (Grant numbers: P500PM_206620 and P5R5PM_230625) and the National Multiple Sclerosis Society (Grant number: FG-2107-38022). UH and FE are supported by the Max Planck Society and the European Research Council [ERC StG 758974]. The Rick Hansen Spinal Cord Injury Registry is supported by funding from the Praxis Spinal Cord Institute through the Government of Canada and the Province of British Columbia. DSR and GN are funded by the Intramural Research Program of the National Institute of Neurological Disorders and Stroke, NIH, USA.

# Supplementary Material: Minimizing morphometric drift in lifelong learning segmentation of the spinal cord

This document contains the supplementary material presenting an overview of the dataset characteristics, additional plots comparing the variability in spinal cord cross-sectional area (CSA) per individual contrasts, CSA variability split across vendors, and quantitative comparison in terms of the Dice scores, relative volume error and average surface distance.

## S1.1. Dataset characteristics

**Table S1** contains a detailed overview of the range of image resolutions and orientations for each contrast in the dataset. Note that the images span a wide range of resolutions, especially with thick slices in the axial orientation for a few contrasts.

***Table S1.*** *Dataset characteristics grouped by image orientation (axial, sagittal) and resolution (isotropic, anisotropic) for each contrast. Mean in-plane resolution and mean slice thickness are shown, followed by their respective minimum and maximum range of resolutions (in square brackets).*

| **Contrasts** | **Isotropic** | | **Anisotropic Axial Orientation** | | **Anisotropic Sagittal Orientation** | |
|---|---|---|---|---|---|---|
| | in-plane resolution ($mm^2$) | slice thickness (mm) | in-plane resolution ($mm^2$) | slice thickness (mm) | in-plane resolution ($mm^2$) | slice thickness (mm) |
| **T1-w** | 1.0 x 1.0 [1.0, 1.0] | 1.0 [1.0, 1.0] | 0.35 x 0.35 [0.35 x 0.35, 0.35 x 0.35] | 2.54 [2.5, 5.0] | 1.0 x 1.0 [1.0, 1.0] | 1.0 [1.0, 1.0] |
| **T2-w** | 0.8 x 0.8 [0.8, 0.8] | 0.8 [0.8, 0.8] | 0.5 x 0.5 [0.3 x 0.3, 0.8 x 1.0] | 3.8 [1.0, 7.0] | 0.48 x 0.48 [0.28 x 0.28, 0.96 x 0.96] | 2.13 [0.8, 4.83] |
| **T2*-w** | – | – | 0.44 x 0.44 [0.29 x 0.29, | 4.93 [2.5, 9.2] | – | – |
| | | | 0.5 x 0.5] | | | |
| **MT-on** | – | – | 0.89 x 0.89 [0.62 x 0.62, 0.9 x 0.9] | 5.06 [5.0, 9.3] | – | – |
| **GRE-T1w** | – | – | 0.89 x 0.89 [0.68 x 0.68, 0.9 x 0.9] | 5.0 [5.0, 5.0] | – | – |
| **DWI** | – | – | 0.89 x 0.89 [0.34 x 0.34, 1.0 x 1.0] | 5.0 [4.91, 5.0] | – | – |
| **PSIR** | – | – | – | – | 0.69 x 0.69 [0.67 x 0.67, 0.69 x 0.69] | 3.0 [3.0, 3.0] |
| **STIR** | – | – | – | – | 0.7 x 0.7 [0.7 x 0.7, 0.7 x 0.7] | 3.0 [3.0, 3.0] |
| **MP2RAGE UNIT1** | 1.0 x 1.0 [1.0, 1.0] | 1.0 [1.0, 1.0] | – | – | | |

**Table S2** contains the dataset characteristics of the frozen test set of healthy participants used to generate CSA plots (Figure 6) and evaluate morphometric variability between various model versions.

***Table S2.*** *Characteristics of the test set of healthy participants*

| | **Values** |
|---|---|
| **Number of participants** | 49 |
| **Sequences** | T1-w, T2-w, T2*-w, MT-on, GRE-T1w, DWI |
| **Number of MRI scans** | 294 (49 * 6 sequences) |
| **Sex** **(Male / Female)** | 22 / 27 |
| **Age (y)** **(mean ± standard deviation)** | 28.5 ± 6.2 |
| **Age range (y)** | 21 – 50 |

| | | | |
|---|---|---|---|
| **MRI manufacturers** | Siemens (n=30), Philips (n=13), GE (n=6) | | |
| **MRI field strength** | 3T (n=49) | | |
| | | **In-plane resolution ($mm^2$)** | **Slice thickness (mm)** |
| **MRI Sequence parameters** | T1-w | 1 × 1 | 1 |
| | T2-w | 0.8 × 0.8 | 0.8 |
| | T2*-w | [0.44–0.5] x [0.44–0.5] | 5.0 |
| | MTon | [0.67–0.9] x [0.67–0.9] | 5.0 |
| | GRE-T1w | [0.68–0.9] x [0.68–0.9] | 5.0 |
| | DWI | [0.34–0.9] x [0.34–0.9] | 5.0 |

## S1.2. Training hyperparameters

In Table S3, we show the list of training hyperparameters and their corresponding values.

***Table S3.*** *Hyperparameters used during training and their corresponding values.*

| Hyperparameter | Values |
|---|---|
| Optimizer | Stochastic Gradient Descent (SGD) |
| Initial learning rate | 0.01 (1e-2) |
| Weight decay | 0.00003 (3e-5) |
| Momentum | 0.99 |
| Learning rate scheduler | PolynomialLR |
| Decay rate | 0.9 |
| Number of epochs | 1000 |
| Iterations per epoch | 250 |

## S1.2. CSA variability across individual contrasts

**Figure S1** shows the variability in the spinal cord CSA across six contrasts on the test set (*n=49*; $n_{vol}$*=294*) of the spine-generic multi-subject database (Cohen-Adad et al., 2021a) between three methods: (i) `sct_deepseg_sc` (Gros et al., 2019), (ii) our previous model, `contrast-agnostic_v2.0` (Bédard et al., 2025), and the proposed model, `contrast-agnostic_v3.0`. Compared to the previous version (`v2.0`), our proposed model (`v3.0`) achieves a similar CSA variability on the test set of healthy participants despite being trained on heterogeneous data containing new contrasts and several pathologies.

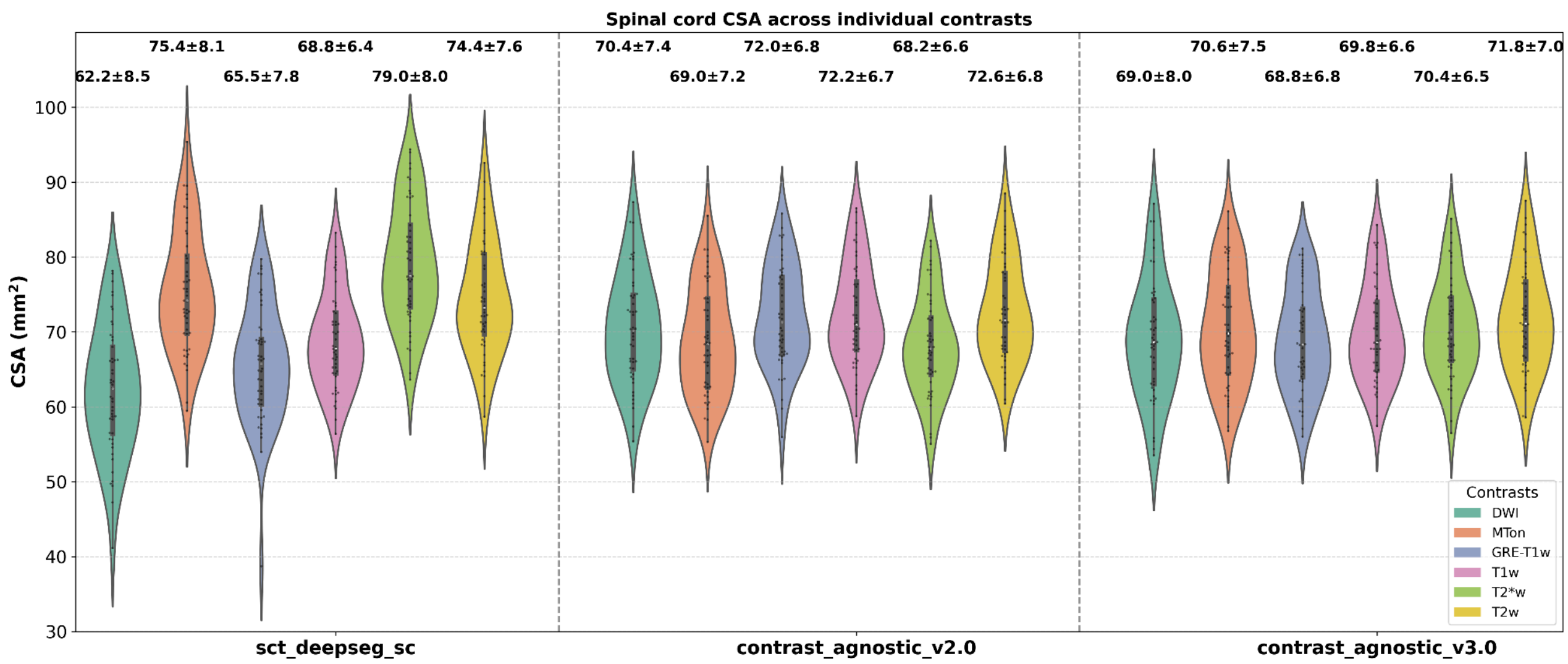

***Figure S1.*** *Variability in spinal cord CSA across 6 contrasts compared with existing automatic segmentation methods on a test set of healthy participants (n=49). Even after the addition of new pathologies and contrasts to the training set, CSA variability achieved by the proposed `contrast-agnostic_v3.0` model remains similar to our previous model `contrast-agnostic_v2.0` (trained only on a healthy participants database) and shows a substantial improvement over `sct_deepseg_sc`.*

## S1.3. CSA variability across scanner manufacturers

In this section, we evaluate the variability in the CSA measurements for a *single participant* across different scanner manufacturers. We used the spine-generic `data-single-subject` dataset (Cohen-Adad et al., 2021), which includes cervical spinal cord scans in a single healthy participant using six contrasts (T2w, T1w, T2*w, MT-on, GRE-T1w, and DWI) across 15 sites with 3 scanner vendors (GE; *n=4*, Philips; *n=4*, Siemens; *n=7*). As with the previous evaluations, we compared three methods: `sct_deepseg_sc` (Gros et al., 2019), `contrast_agnostic_v2.0` (Bédard et al., 2025), and the proposed `contrast_agnostic_v3.0`, for contrasts and sites. In all comparisons, the spinal cord segmentations were obtained independently for each of the above methods, and the vertebral levels were identified using `sct_label_vertebrae`. Then, we calculated the CSA averaged across C2-C3 vertebral levels and computed its standard deviation (STD) across scanner manufacturers.

It is important to stress that all data points represent the *same* participant. each of the 6 contrasts comparing the two segmentation methods across all 15 sites. **Figure S2** presents the CSA STD across 6 contrasts per site for both segmentation methods, separated per MRI vendor. The STD using the `contrast_agnostic_v3.0` method yields a lower STD than when using `sct_deepseg_sc` for segmentation, and is very similar to `contrast_agnostic_v2.0`.

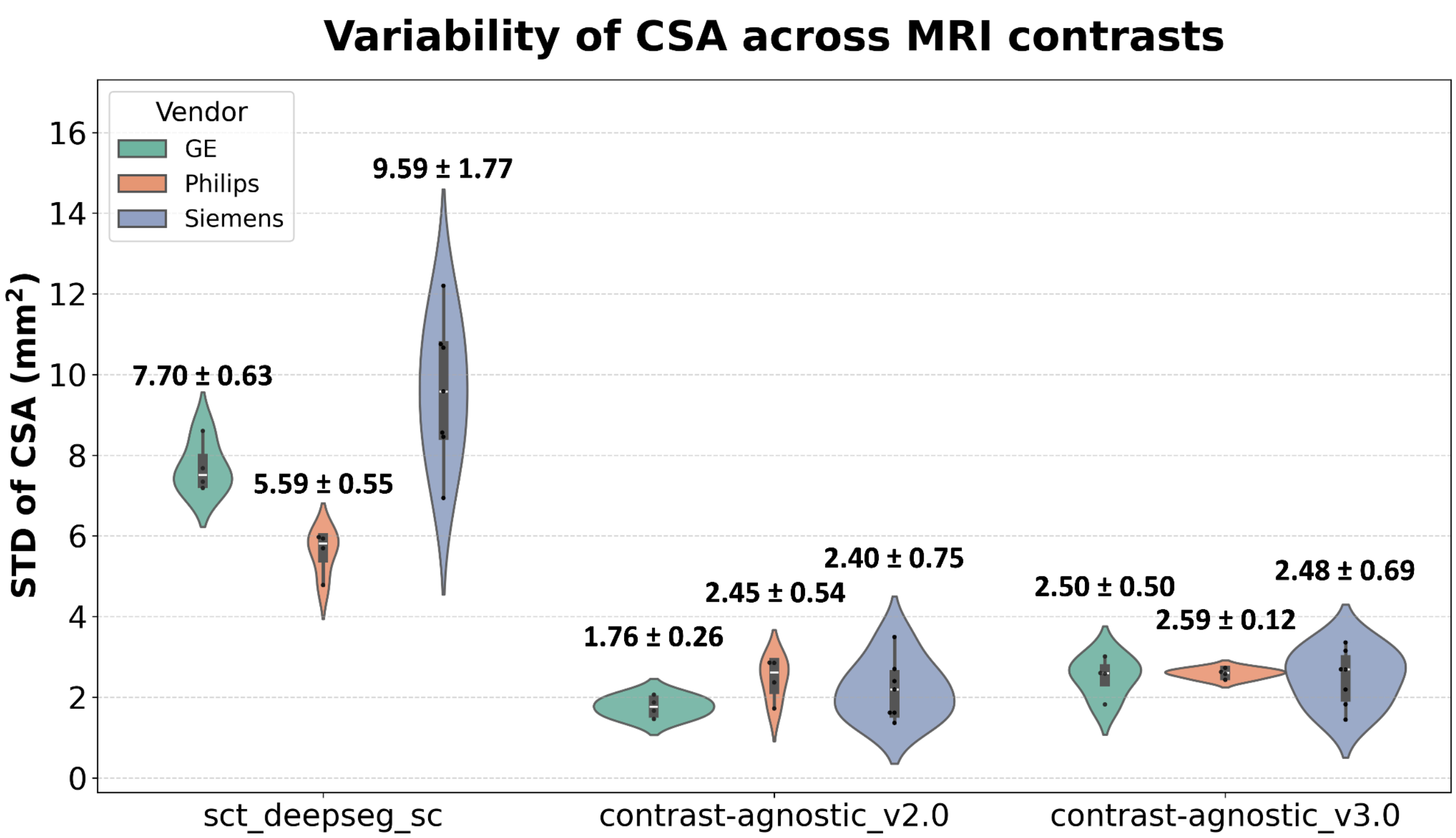

***Figure S2.*** *Variability of spinal cord CSA across contrasts separated per vendor for segmentations generated with sct_deepseg_sc (Gros et al., 2019), contrast-agnostic_v2.0 (Bédard et al., 2025) and contrast-agnostic_v3.0 (proposed) segmentation and contrast-agnostic of the same participant scanned across 15 different MRI sites. Each dot represents one site; mean and standard deviation are presented above.*

## S1.4. CSA variability with recursively generated labels

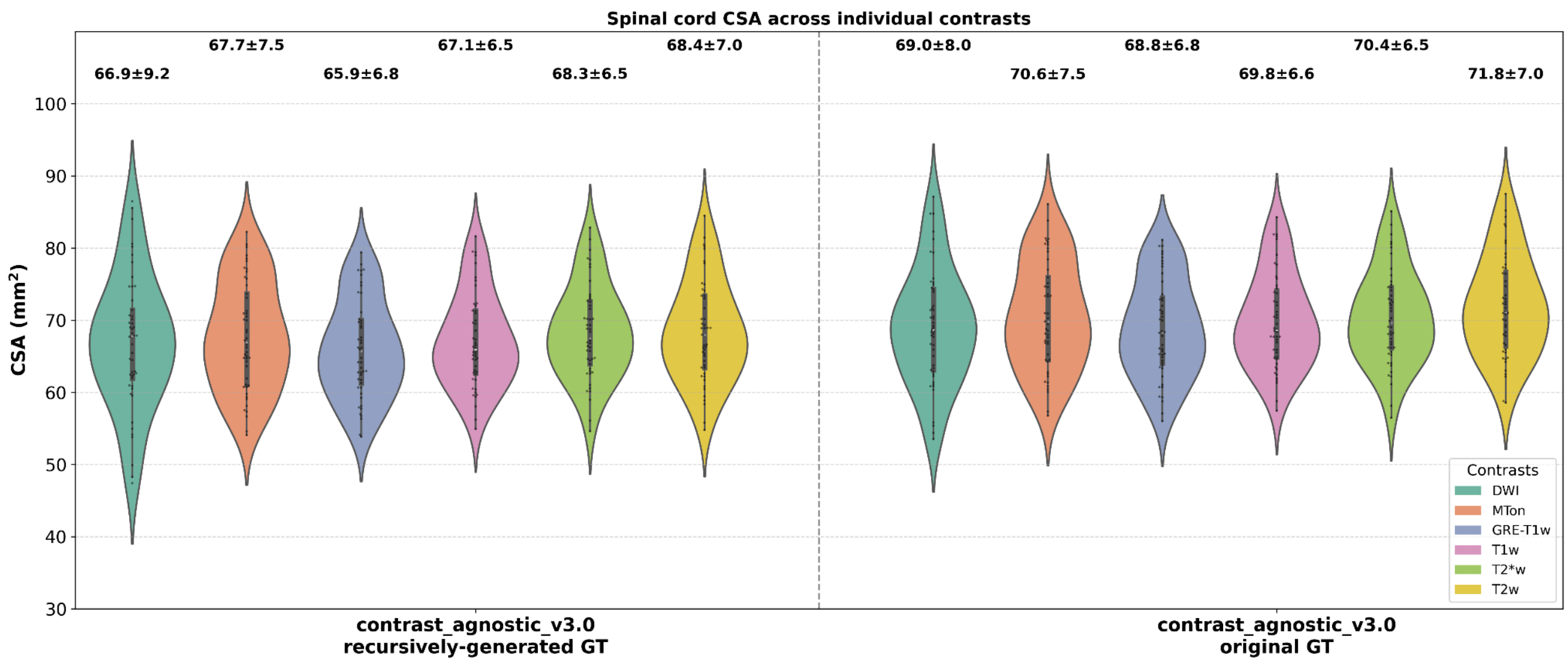

**Figure S3.** *Variability in spinal cord CSA across 6 contrasts on a test set of healthy participants (n=49) compared between the models trained with the: (i) original distribution of GT masks created from a mix of manual annotations and automatic segmentation methods, and (ii) GT masks regenerated with contrast_agnostic_v3.0 model without any manual corrections. The model trained on recursively generated GT masks achieved lower average CSA per contrast compared to the model trained on the original distribution of GT masks on all contrasts.*

**Figure S3** plots the average CSA per contrast for the ablation study, comparing the downstream effect of training the contrast_agnostic_v3.0 model on the original distribution of GT masks and the masks generated recursively without any manual correction.